%% file: iclr2026_conference.tex
\documentclass{article} 
\usepackage{iclr2026_conference,times}
\usepackage{amssymb}
\usepackage{times}
\usepackage{multicol}
\usepackage{epsfig}
\usepackage{graphicx}
\usepackage{booktabs}
\usepackage{color}
\usepackage{amsmath,bm}
\usepackage{multirow}
\usepackage{float}
\usepackage{subcaption} 
\usepackage{tabularray}
\usepackage{array}
\usepackage{pifont}
\usepackage{enumitem}
\usepackage{colortbl}
\usepackage{tabularx} 

\usepackage{algorithm, algpseudocode} 
\usepackage{listings} 
\usepackage{makecell}
\usepackage{float} 
\usepackage{comment}
\usepackage{adjustbox}
\usepackage{multirow}
\usepackage{wrapfig}
\input{math_commands.tex}

\usepackage[bookmarks=true,pagebackref,breaklinks,colorlinks]{hyperref}
\usepackage{url}

\input{math_commands}
\lstdefinelanguage{Python}{
    keywords={while, for, if, else, elif, def, return, import, from, as, break, continue, pass},
    keywordstyle=\color{blue},
    comment=[l]{\#},
    commentstyle=\color{gray},
    string=[s]{"}{"},
    stringstyle=\color{red},
    morecomment=[s]{'''}{'''},
    morecomment=[s]{"""}{"""},
}

\lstdefinestyle{python}{
    basicstyle=\ttfamily\small,
    breaklines=true,
    frame=none,
    numbers=none,
    numberstyle=\tiny\color{gray},
    language=Python, 
}

\title{\textbf{\emph{OpenFly}}: A Comprehensive Platform for Aerial Vision-Language Navigation}

\author{
\textbf{Yunpeng Gao$^{1,2}$\textsuperscript{*}, Chenhui Li$^{1}$\textsuperscript{*}, Zhongrui You$^{1,3}$\textsuperscript{*}, Junli Liu$^{1,2}$\textsuperscript{*}, Zhen Li$^{1,4}$\textsuperscript{*}}, \\
\textbf{ Pengan Chen$^{1,5}$, Qizhi Chen$^{1,6}$,Zhonghan Tang$^{1,7}$, Liansheng Wang$^{1}$, Penghui Yang$^{1,8}$},  \\
\textbf{Yiwen Tang$^{1,2}$, Yuhang Tang$^{1,2}$, Shuai Liang$^{1,9}$, Songyi Zhu$^{1}$, Ziqin Xiong$^{1,9}$, Yifei Su$^{1,10}$}, \\
\textbf{Xinyi Ye$^{1}$, Jianan Li$^{1}$, Yan Ding$^{1}$, 
Dong Wang$^{1}$, Xuelong Li$^{11}$, {Zhigang Wang}$^{1}$\textsuperscript{\dag}, Bin Zhao$^{1,2}$\textsuperscript{\dag} }
\\ \\
$^1$Shanghai AI Laboratory, 
$^2$Northwestern Polytechnical University, \\
$^3$Beihang University, 
$^4$Shanghai Jiao Tong University, \\
$^5$The University of Hong Kong,
$^6$Zhejiang University,  \\
$^7$University of Science and Technology of China, \\
$^8$East China University of Science and Technology, 
$^9$Fudan University, \\
$^{10}$Institute of Automation, Chinese Academy of Sciences,
$^{11}$TeleAI 
\\
\\
}

\iclrfinalcopy 
\begin{document}

\maketitle

\renewcommand{\thefootnote}{\fnsymbol{footnote}}
\footnotetext[1]{Equal Contribution.}
\footnotetext[2]{Corresponding Author.}

{%
\renewcommand\twocolumn[1][]{#1}
\maketitle
\begin{center}
\centering
\begin{minipage}[t]{\linewidth}
\vspace{-0.8cm}
{\captionsetup{type=figure}  
\includegraphics[width=\textwidth]{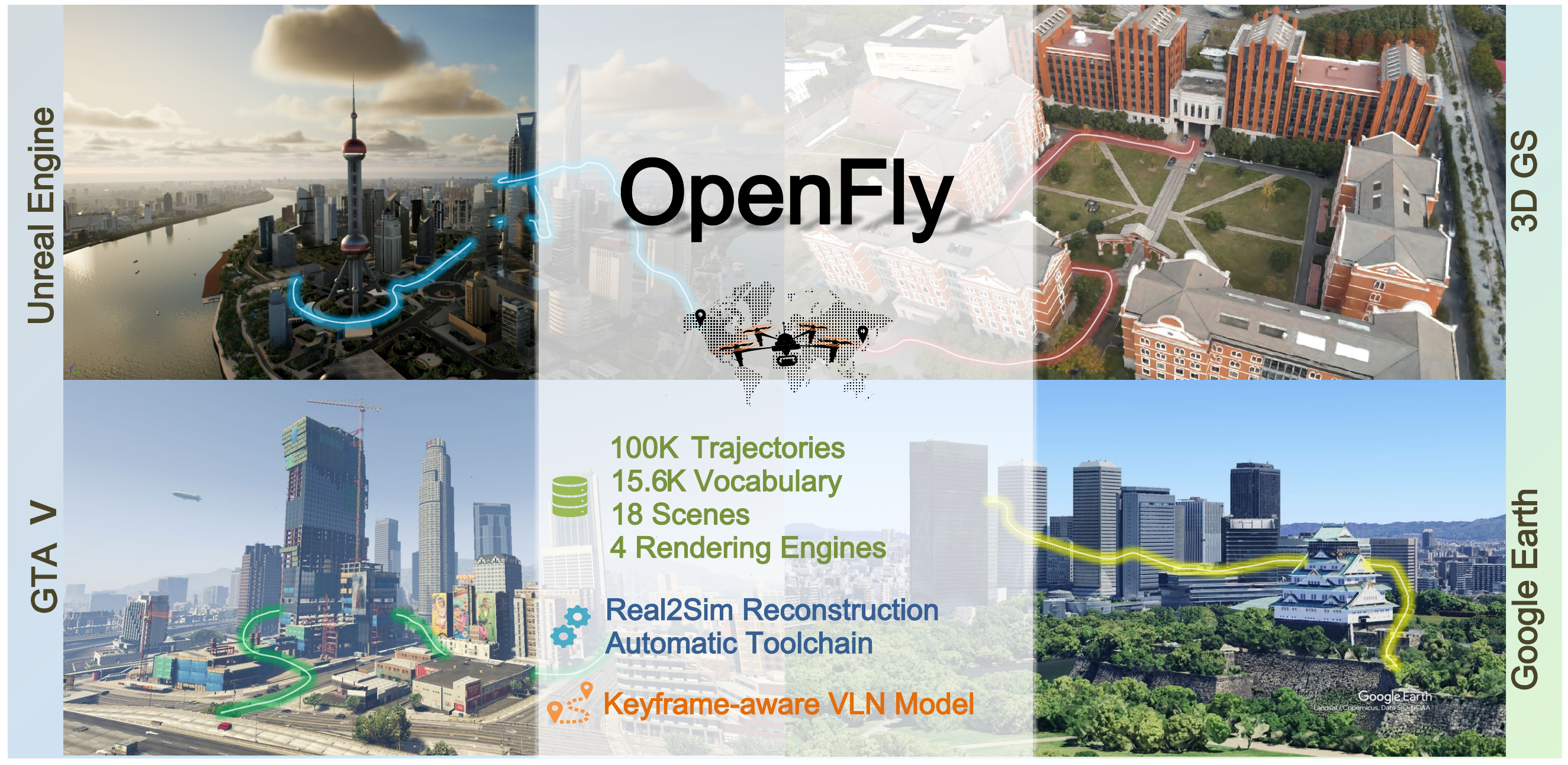}
\captionof{figure}{\footnotesize{{
Overview of OpenFly. This work consists of (1) the integration of 4 rendering engines, significantly enhancing the diversity of scenario resources for aerial vision-language navigation; (2) an automatic data generation toolchain, eliminating reliance on labor-intensive annotations; (3) the largest aerial VLN dataset to date, comprising 100K trajectories; and (4) a keyframe-aware VLN model, achieving superior performance in both simulated and real-world scenes.
}}}
\label{fig:OpenFly}}
\end{minipage}
\end{center}
}

\begin{abstract}
Aerial Vision-Language Navigation (VLN) seeks to guide UAVs by leveraging language instructions and visual cues, establishing a new paradigm for human-UAV interaction. However, the collection of VLN data demands extensive human effort to construct trajectories and corresponding instructions, hindering the development of large-scale datasets and capable models. To address this problem, we propose \textbf{OpenFly}, a comprehensive platform for aerial VLN. Firstly, OpenFly integrates 4 rendering engines and advanced techniques for diverse environment simulation, including Unreal Engine, GTA V, Google Earth, and 3D Gaussian Splatting (3D GS). Particularly, 3D GS supports real-to-sim rendering, further enhancing the realism of our environments. Secondly, we develop a highly automated toolchain for aerial VLN data collection, streamlining point cloud acquisition, scene semantic segmentation, flight trajectory creation, and instruction generation. Thirdly, based on the toolchain, we construct a large-scale aerial VLN dataset with 100k trajectories, covering samples of diverse scenarios and assets across 18 scenes. Moreover, we propose OpenFly-Agent, a keyframe-aware VLN model emphasizing key observations to promote performance and reduce computations. For benchmarking, extensive experiments and analyses are conducted, where our navigation success rate outperforms others by 14.0\% and 7.9\% on the seen and unseen scenarios, respectively. The toolchain, dataset, and codes will be open-sourced. 
\end{abstract}

\input{sec/1_introduction}
\input{sec/2_related_work}
\input{sec/3_Dataset}

\input{sec/4_Methodology}
\input{sec/5_Experiments}
\input{sec/6_Conclusion}

\section{Acknowledgement}
This work is supported by Shanghai AI Laboratory.

\bibliography{BibForOpenFly}
\bibliographystyle{iclr2026_conference}

\clearpage

\appendix
\section*{Appendix}
\input{sec/supp}

\end{document}

%% file: math_commands.tex
\usepackage{amsmath,amsfonts,bm}

\def\eqref#1{equation~\ref{#1}}

\def\1{\bm{1}}

\DeclareMathAlphabet{\mathsfit}{\encodingdefault}{\sfdefault}{m}{sl}
\SetMathAlphabet{\mathsfit}{bold}{\encodingdefault}{\sfdefault}{bx}{n}



%% file: sec/1_introduction.tex
\section{Introduction} ~\label{Sec:Introduction}
\begin{figure*}
    \centering
    \includegraphics[width=0.99\linewidth]{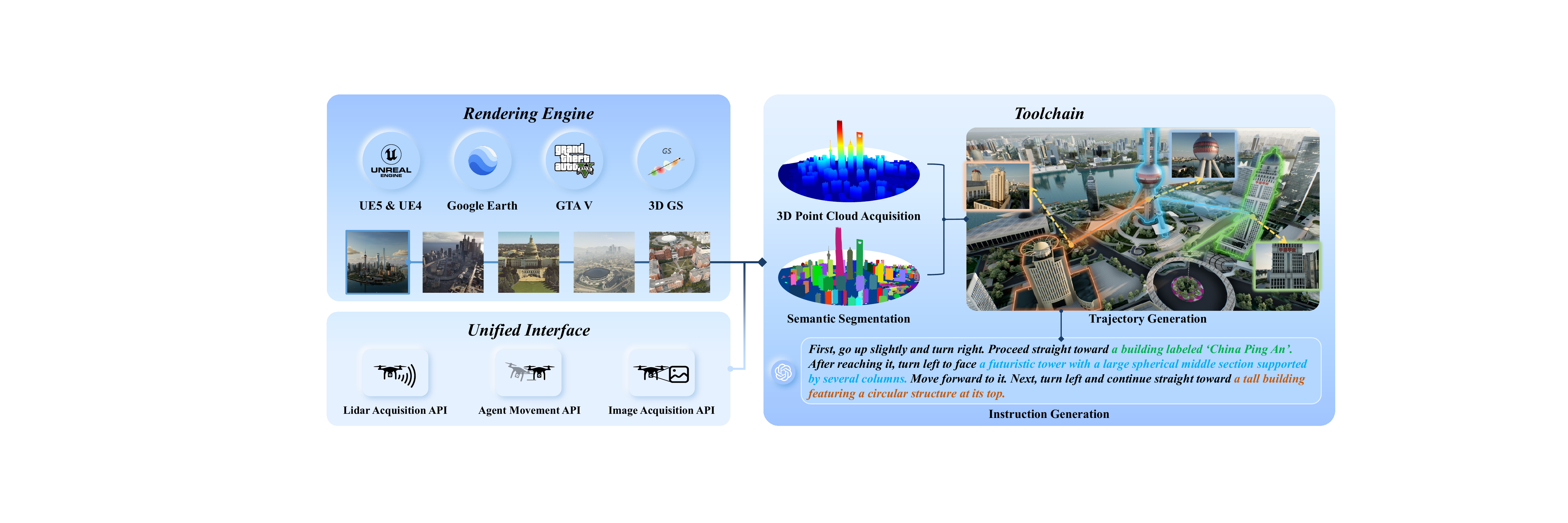}
    \caption{Framework of the automatic data generation. Multiple rendering engines are integrated to provide diverse, high-quality scenes. Built on these, several interfaces and tools are developed to enable automated generation of trajectories and instructions.}
    \label{fig:data_gen}
    \vspace{-0.1cm}
\end{figure*}
Embodied AI has drawn growing research attention, where vision-language navigation (VLN) emerging as a core task that navigate agents to a target location according to linguistic instructions and visual observations. A number of benchmark datasets have been established, \emph{e.g.,} TouchDown~\citep{Touchdown}, REVERIE~\citep{REVERIE}, R2R~\citep{R2R}, RxR~\citep{RxR}, CVDN~\citep{CVDN}, VLN-CE~\citep{VLN-CE}, and LANI~\citep{LANI}, which have significantly advanced the development of VLN methods ~\citep{instructnav, CMA, ETPNav, MGMap, navid, cognav, iglnav, ma2025}. Nevertheless, existing efforts primarily target indoor or ground-based agents, while unmanned aerial vehicles (UAVs), crucial for aerial photography, rescue operations, and cargo transport, remain unexplored.


Most recently, AerialVLN~\citep{aerialVLN} and OpenUAV~\citep{openuav} have made significant strides by leveraging UAV simulators to mitigate the scarcity of aerial VLN datasets, thereby driving advances in this field. However, several critical challenges remain to be addressed:

\begin{itemize}[left=0pt]
\item \textbf{Limited data diversity.} Existing methods rely on AirSim and Unreal Engine (UE) for UAV control, which confines them to digital assets compatible with these platforms, limiting the diversity of available data and constraining the incorporation of more photorealistic sources.

\item \textbf{High collection cost.} The process of generating trajectories relies on pilots operating UAVs in simulators, followed by manual annotation to create language instructions. The entire process is labor-intensive, time-consuming, and difficult to scale.

\item  \textbf{Small data scale.} Current datasets for aerial VLN remain relatively small, containing only about 10k trajectories, which is far behind embodied manipulation datasets. By contrast, Open X-Embodiment~\citep{open_x-embodiment} and  EO-1~\citep{eo1} have collected over 1M episodes of manipulation, significantly promoting the development of vision-language-action (VLA) models. 
\end{itemize} 

To address these issues, we propose \textbf{OpenFly}, a comprehensive platform consisting of diverse rendering engines, a versatile toolchain, and a large-scale benchmark for the aerial VLN task. \textbf{To enhance data diversity}, the platform is established on various widely-used rendering engines and advanced techniques, \emph{i.e.,} UE, GTA V, Google Earth, and 3D Gaussian Splatting (3D GS), enabling us to utilize a wide range of assets as shown in Fig. \ref{fig:OpenFly}. In particular, we use UAVs to capture numerous real-world images and integrate 3D GS technology into our platform to reconstruct realistic 3D scenes, empowering real-to-sim simulation. \textbf{To improve the efficiency of data collection}, we develop a versatile toolchain for automated aerial VLN data generation as depicted in Fig. \ref{fig:data_gen}. Specifically, point cloud acquisition is first conducted to capture the 3D occupancy of a scene. Next, scene semantic segmentation is performed to identify and select landmarks as waypoints along the flight trajectories. Building on these tools, trajectory generation is then carried out, taking landmarks and point clouds as input, using predefined flight actions as basic units, and automatically searching for a collision-free trajectory. Finally, we feed the trajectories and corresponding UAV-egocentric images into a vision-language-model (VLM), \emph{e.g.,} GPT-4o, to generate linguistic instructions. The entire pipeline is highly automated, reducing the reliance on UAV pilots and annotators. \textbf{To collect a large-scale dataset}, we meticulously collected 18 high-quality scenes, generating various trajectories of differing heights and lengths. Benefitting from our toolchain, we are able to quickly construct a dataset of \textbf{100k} samples, significantly larger than existing datasets.

Besides, we propose \textbf{OpenFly-Agent}, a keyframe-aware aerial VLN model incorporating an adaptive frame-level sampling mechanism to emphasize critical observations containing instruction-related landmarks, leading to performance improvement and computation reduction compared to a uniform sampling strategy. Extensive experiments are conducted on the OpenFly dataset to evaluate numerous methods, establishing a comprehensive benchmark for the aerial VLN tasks.
Overall, our contributions can be summarized as follows:

\begin{itemize}[left=0pt]
\item  We build OpenFly on multiple rendering engines and develop a versatile toolchain, enabling the automatic generation of data with high  diversity and efficiency.

\item  We have constructed a large-scale aerial VLN benchmark comprising 100k trajectories across 18 high-quality scenes. To the best of our knowledge, this is the largest aerial VLN benchmark to date, and users can collect more customized data using the OpenFly platform.

\item  We propose OpenFly-Agent, a keyframe-aware VLN model. Extensive experiments in both simulated and real-world settings demonstrate its superior performance. 
\end{itemize}

%% file: sec/2_related_work.tex
\section{Related Works} 
~\label{Sec:related_works}
\subsection{Vision-Language Navigation Datasets}
Numerous datasets have been constructed to accelerate the VLN task. R2R~\citep{R2R} focuses on evaluating agents in unseen buildings and provides discrete navigation options. RxR~\citep{RxR} provides a more densely annotated VLN dataset. TouchDown~\citep{Touchdown} and REVERIE~\citep{REVERIE} have each contributed a dataset from real-life environments, which requires a ground-based agent to follow instructions and find a target object. CVDN~\citep{CVDN} presents a cooperative VLN dataset where agents can access the history of human cooperation for inference. All the above datasets are graph-based, where navigable points are predefined. LANI~\citep{LANI} and VLN-CE~\citep{VLN-CE} propose the VLN task in continuous outdoor/indoor environments, enabling agents to move freely to any unobstructed point. Recently, a few works have tried to construct VLN datasets for aerial space. ANDH~\citep{ANDH} establishes a dialogue-based aerial VLN dataset with bird-view images. CityNav~\citep{CityNav} builds on the point cloud data from SensatUrban~\citep{sensaturban} and linguistic annotations from CityRefer~\citep{CityRefer}, which requires a real-world 2D map to help locate specific landmarks in the instruction. AerialVLN~\citep{aerialVLN}, OpenUAV~\citep{openuav} and CityNavAgent~\citep{citynavagent} integrate AirSim and UE to create VLN scenes where pilots can control UAVs to generate various trajectories.

\subsection{Vision-Language Navigation Methods}

VLN methods enable agents to follow language instructions based on visual observations. Early approaches, such as graph-based methods~\citep{Xiong_2019, Zhang_2019, Srinivasa_2019, Wang_2020}, model the environment as a set of predefined nodes, with agents navigating between these discrete states. However, these methods are limited in dynamic, real-world environments. In recent years, LLM-driven approaches~\citep{Navgpt, NavGPT-2, mapgpt, zeng2025FSDrive} have utilized large language models to enhance reasoning and infer navigation steps, offering more flexibility in continuous environments. Despite significant progress, LLM-based methods still face challenges in grounding language instructions with real-world sensory data and adapting to unknown environments. Meanwhile, training-free LLM-based methods~\citep{hu2025seepointflylearningfree,xu2025geonavempoweringmllmsexplicit} provide a flexible way to infer navigation steps from language alone, enabling rapid adaptation of agents without retraining. In contrast, works like~\citep{Kumar_2021, Maksymets_2021, navid, song2025towards} have shifted focus to continuous spaces, aiming for more realistic navigation in dynamic settings. More recently, aerial VLN has gained attention, with AerialVLN~\citep{aerialVLN} proposing a lookahead guidance method for better training trajectories, while STMR~\citep{stmr} enhances spatial reasoning through matrix representations, and OpenUAV~\citep{openuav} integrates human feedback with ground-truth trajectories to guide navigation.

%% file: sec/3_Dataset.tex
\section{Automatic Data Generation}
\label{sec:Automatic}

In this section, we first introduce the rendering engines and data resources, then present the developed toolchain. The overall framework for automatic data generation is illustrated in Fig. \ref{fig:data_gen}.

\subsection{Rendering Engines and Data Resources}
We leverage multiple rendering engines to construct diverse and realistic environments. Specifically, \textbf{Unreal Engine} provides eight urban scenes spanning over $100km^2$ with rich assets such as buildings, vehicles, and pedestrians. \textbf{GTA V} contributes a highly realistic cityscape modeled after Los Angeles. \textbf{Google Earth} offers four urban regions (Berkeley, Osaka, Washington D.C., and St. Louis) covering $53.60km^2$. Besides, hierarchical \textbf{3D Gaussian Splatting} ~\citep{kerbl2024hierarchical} is employed for the reconstruction of real-world environments from UAV data, encompassing more than $7km^2$ across five campuses with diverse landmarks. More details and examples are provided in Appendix~\ref{appen_rendering_engines}.

\subsection{Toolchain for Automatic Data Collection}
\label{toolchain}
\indent \indent To achieve automatic data generation, we first integrate the above rendering engines and design three unified interfaces to control the agent movement and acquire sensor data (presented in Appendix \ref{appen_interface}). Based on these interfaces, we further develop a toolchain, streamlining point cloud acquisition, scene semantic segmentation, trajectory creation, and instruction generation.

\textbf{3D Point Cloud Acquisition.} 
OpenFly integrates various rendering engines and scenes, exhibiting distinct characteristics. To address these differences, we provide two methods to reconstruct the point cloud map for different scenes. 1) Rasterized Sampling Reconstruction:
For UE and GTA V scenes, we customize rasterized sampling points at appropriate resolutions, followed by using the developed interface to obtain the local point cloud at the sampling points and stitch them for the entire scene. 2) Image-Based Sparse Reconstruction: In 3D GS, the scene reconstruction process begins with the open-source COLMAP~\citep{colmap} framework, which generates a sparse point cloud from input images. We directly export and use the point clouds from this step.

\textbf{Scene Semantic Segmentation.} 
VLN requires meaningful landmarks as navigation targets. Thus, we offer three semantic segmentation methods to identify landmarks. 1) 3D Scene Understanding: A sequence of top-down views of the scene is captured in a rasterized format, followed by the off-the-shelf Octree-Graph~\citep{octree_graph} to extract semantic 3D instances. 2) Point Cloud Projection and Contour Extraction: We acquire the point cloud of a scene and project the voxelized point cloud onto the ground. For each instance, its contour is segmented, and the maximum height of its points is used as the final height. Additionally, semantic annotations are obtained by feeding the segmented instances to GPT-4o for caption. 3) Manual Annotation: When the point cloud quality of a scene is low or finer segmentation is required, OpenFly provides an interface for manually annotating instances and semantics within the point cloud. Users can choose these methods flexibly based on their requirements. The corresponding details and results are shown in Appendix \ref{appen_pc_ss}.

\textbf{Automatic Trajectory Generation.}
Leveraging the point cloud map and segmentation tools, OpenFly can automatically generate VLN trajectories using the following method. First, a global voxel map $M_{global}$ is constructed from the scene point cloud. Second, a landmark is randomly chosen as the target, with a starting point being selected at a certain distance from the landmark, and a point close to the landmark being chosen as the endpoint. Third, A collision-free trajectory is generated using the A*~\citep{astar} pathfinding algorithm based on $M_{global}$ and a customized action space. By repeatedly selecting the endpoint as the new starting point, complex trajectories can be generated. Finally, utilizing OpenFly's interface, UAV-egocentric images corresponding to the trajectory points are obtained as visual observations. More details are included in Appendix \ref{appen_trajectory}.

\textbf{Automatic Instruction Generation.}
Most previous works have predominantly relied on manual annotation to generate  instructions, which is costly and hinders dataset scalability~\citep{aerialVLN,CityNav,Liu_2025_ICCV}. To address this issue, we propose a highly automated instruction generation method based on VLMs, \emph{e.g.,} GPT-4o.

A straightforward method would be to input all images to VLMs to analyze the trajectory and generate instructions. However, using all images introduces considerable computational overhead and causes significant difficulties for a VLM to understand. Additionally, we find the `Forward' action usually occupies a larger proportion of a flight trajectory, with `Turn Left/Turn Right' or `Ascend/Descend' actions taken when encountering key landmarks. Therefore, we split the complete trajectory into multiple sub-trajectories according to action transitions, extracting key actions and images for processing. Notably, slight angle adjustments often occur during flight to change direction subtly, which will be ignored in this procedure. We submit the action sequence and the last captured three images of each sub-trajectory to a VLM to generate a sub-instruction of both the action and the landmark.
All sub-instructions of the same trajectory are then processed by an LLM to integrate into a complete instruction. The proposed strategy significantly improves the instruction accuracy compared to directly inputting all trajectory images to a VLM. To further verify the data quality, we randomly select 3K samples from the entire dataset according to the data distribution in Sec.~\ref{data_split}. After manually inspecting these samples, we find that they reach a high qualification rate of 91\%. The problematic data involves some vague descriptions, but it is still considered acceptable by examiners. Besides, all test data have undergone manual inspection, with low-quality ones removed. Thanks to GPT's high concurrency, we can quickly generate a large number of instructions, which solves the problem of difficult and time-consuming manual annotation. \textbf{More details of instruction generation and data quality control are provided in Appendix \ref{appen_instruction} and \ref{appen_quality}.}


\begin{table}[t]
\small      
\caption{Comparisons of different VLN datasets. $N_{traj}$: the number of total trajectories. $N_{vocab}$: vocabulary size. Path Len: the average length of trajectories, measured in meters. Intr Len: the average length of instructions. $N_{act}$: the average number of actions per trajectory.}
\begin{adjustbox}{center}
\renewcommand{\arraystretch}{1.0}
\resizebox{\textwidth}{!}{ 
\begin{tabular}{lcccclcc}
\toprule
Dataset   & $N_{traj}$ & $N_{vocab}$ & Path Len. & Intr Len. & Action Space & $N_{act}$ & Environment \\ \midrule
R2R~\citep{R2R}       & 7189      & 3.1K         & 10.0      & 29        &graph-based   & 5       & Matterport3D  \\
RxR~\citep{RxR}       & 13992     & 7.0K         & 14.9      & 129       &graph-based   & 8       & Matterport3D  \\
TouchDown~\citep{Touchdown} & 9326      & 5.0K         & 313.9     & 90        &graph-based   & 35      & Google Street View  \\ 
VLN-CE~\citep{VLN-CE}    & 4475      & 4.3K         & 11.1      & 19        &2 DoF         & 56      & Matterport3D   \\
AerialVLN~\citep{aerialVLN} & 8446      & 4.5K         & 661.8     & 83        &4 DoF         & 204     & AirSim + UE \\
CityNav~\citep{CityNav}   & 32637     & 6.6K         & 545       & 26        &4 DoF         & -       & SensatUrban  \\
OpenUAV~\citep{openuav}   &12149      &10.8K         & 255       & 104       &6 DoF         & 264     & AirSim + UE \\ \midrule
\multirow{3}{*}{Ours} &\multirow{3}{*}{100K}     &\multirow{3}{*}{15.6K}        &\multirow{3}{*}{99.1}        &\multirow{3}{*}{59}       &\multirow{3}{*}{4 DoF}    &\multirow{3}{*}{35}     & \parbox[t]{3cm}{AirSim + UE, GTA V, \\ Google Earth Studio, \\3D GS + SIBR viewers} \\

\bottomrule
\end{tabular}
}
\end{adjustbox}
\label{tab:dataset_comp}
\end{table}

\section{Dataset Analysis}

Table \ref{tab:dataset_comp} summarizes key statistics of several commonly used VLN datasets, from which we can see that our dataset features a significantly larger number of trajectories and a greater environmental diversity. In contrast, our average trajectory length and instruction length are relatively short. This is intentional, we argue that short- and medium-range instructions better reflect natural human usage habits and may be more beneficial for advancing aerial VLN.



\subsection{Trajectory and Instruction Analysis}

\begin{figure}[htbp]
    \centering
    \begin{subfigure}[b]{0.27\textwidth}
        \centering
        \includegraphics[width=\textwidth]{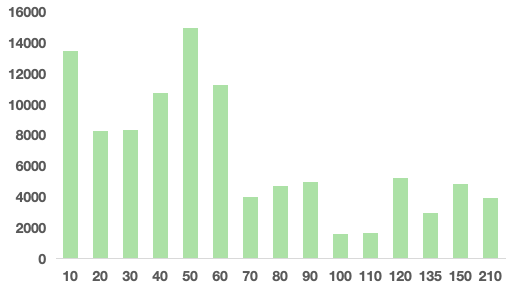}
        \caption{}
        \label{fig:height}
    \end{subfigure}
    \hfill
    \begin{subfigure}[b]{0.28\textwidth}
        \centering
        \includegraphics[width=\textwidth]{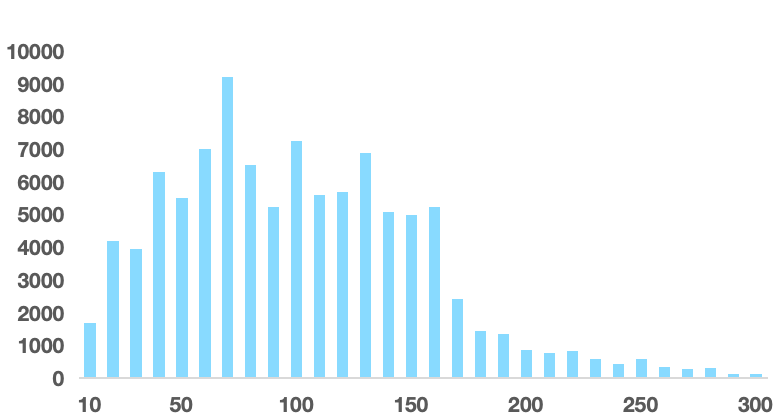}
        \caption{}
        \label{fig:length}
    \end{subfigure}
    \hfill
    \begin{subfigure}[b]{0.21\textwidth}
        \centering
        \includegraphics[width=\textwidth]{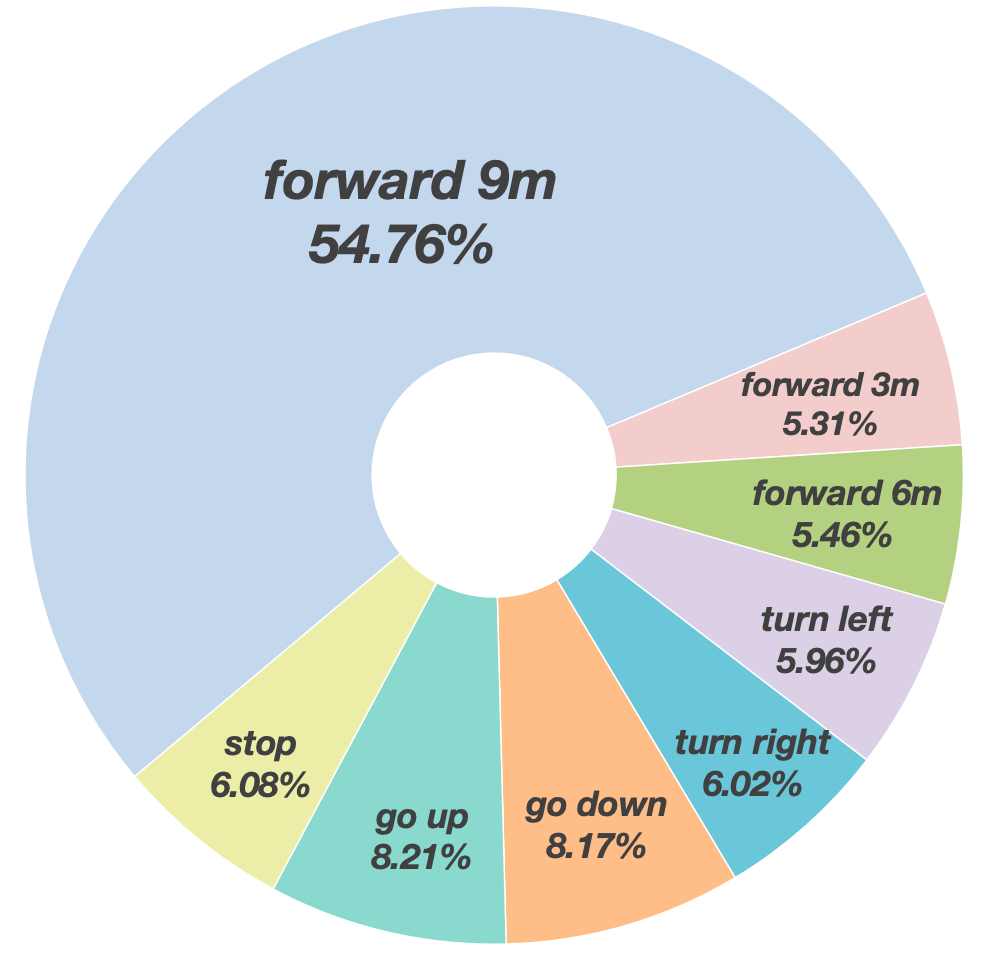}
        \caption{}
        \label{fig:action_distri}
    \end{subfigure}
    \hfill
    \begin{subfigure}[b]{0.2\textwidth}
        \centering
        \includegraphics[width=\textwidth]{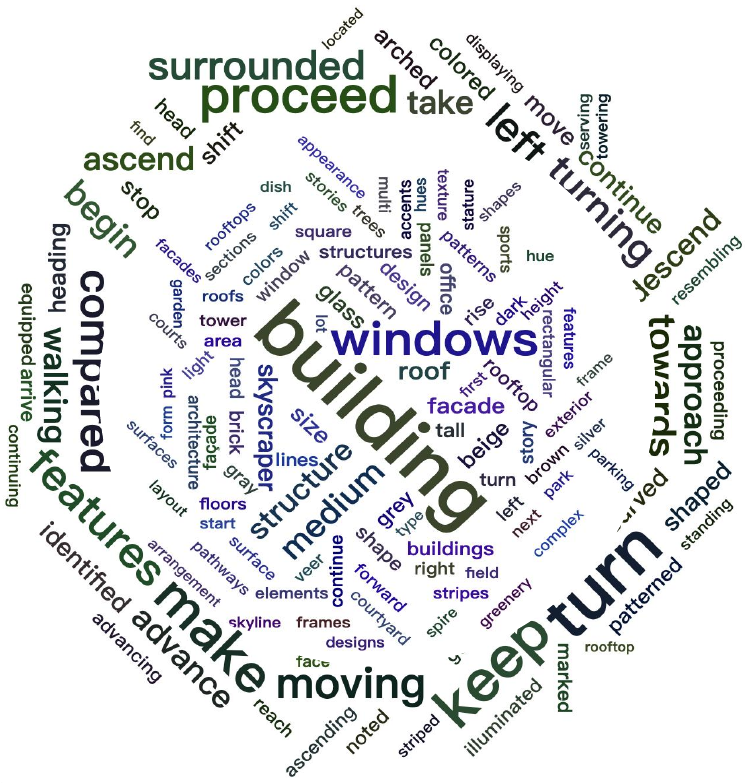}
        \caption{}
        \label{fig:word_cloud}
    \end{subfigure}
    
    \caption{Statistical analysis of the generated data. (a) Height distributions of trajectories. (b) Length distributions of trajectories. (c) Action distributions. (d) Word cloud of verbs and nouns.}
    \label{fig:traj_sta}
\end{figure}

Using our toolchain, we collect a dataset of 100K trajectories, which is much larger than other aerial VLN datasets. Compared with ground-based VLN, the aerial VLN task has more motion dimensions. Therefore, we set different trajectory lengths and flight heights to obtain rich data. Fig. \ref{fig:height} and \ref{fig:length} exhibit the distribution of these data, with their lengths ranging from 0 to 300 meters, and the heights ranging from 0 to 210 meters. Notably, we follow the mainstream methods \citep{VLN-CE, aerialVLN} to use discrete actions, \emph{e.g.,} `Forward' and `Turn left', for trajectory generation, where the step size of the `Forward' action is set to 3 m, 6 m, and 9 m to adapt to targets at different distances. Fig. \ref{fig:action_distri} presents the action distribution of our dataset. It should be noted that collected trajectories also provide corresponding waypoint information, which can be further processed into smoother trajectories to enable navigation waypoint prediction. Besides, the OpenFly platform supports trajectory generation with continuous waypoints directly based on drone trajectory planning algorithms~\cite{WangZXG22, EGO, MellingerK11, ZhouGWLS19}. To further enhance data diversity, we incorporate the DAgger~\cite{DAgger} algorithm as a data augmentation functionality. In summary, OpenFly offers a comprehensive platform that allows users to generate custom data on their own. It also supports agent interaction and enables real-time retrieval of both agent poses and environmental data. This makes it compatible with On-policy training approaches.

For instruction analysis, the vocabulary size of our dataset is 15.6K, and the average length of instructions is 59.
Fig. \ref{fig:word_cloud} illustrates the word clouds of nouns and verbs, where `building', `windows', and `skyscraper' are the most common references, and `proceed' and `turn' are the mostly used verbs for VLN. Due to the space limitation, \textbf{we put more details in Appendix \ref{appen_dataset_analysis}. }

\subsection{Dataset Split}
\label{data_split}
Similar to previous works, we divide the dataset into three splits, \emph{i.e.,}  \textit{Train, Test Seen, Test Unseen}. For the \textit{Train} split, 7 scenes under the UE rendering engine account for $75.7\%$ of all data, since UE provides the largest number of scenes, where different amounts of trajectories are sampled according to the areas of scenes. The 4 scenes created by 3D GS are also the main part of the data, accounting for nearly $20\%$ of the total amount. To ensure visual quality, we only collect data from a high-altitude perspective using Google Earth, which accounts for $4.46\%$. The \textit{Test Seen} data consists of 1800 trajectories uniformly sampled from 11 seen scenarios, and the \textit{Test Unseen} data comprises 1200 trajectories uniformly generated from 3 unseen scenes, \emph{i.e.,} UE-smallcity, 3D GS-sjtu02, and a Los Angeles-like city in GTA V. Detailed data distributions are shown in Appendix \ref{appen_dataset_analysis}.

%% file: sec/4_Methodology.tex
\section{OpenFly-Agent}

\begin{figure*}[tp]
\centering
    \includegraphics[width=0.98\linewidth]{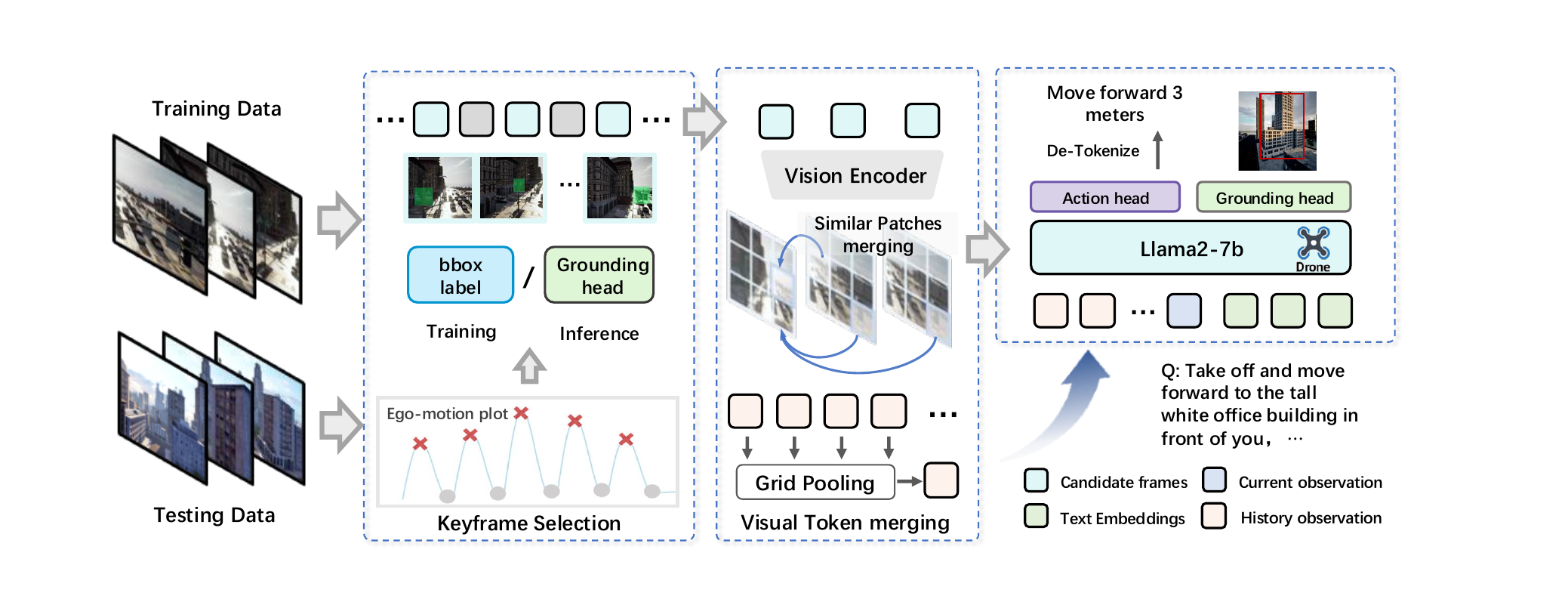}
    \caption{The architecture of OpenFly-Agent. Keyframes are selected according to action transitions and the landmark grounding module to extract crucial observations as the history, with corresponding visual tokens compressed to further reduce the computational burden.}
    \label{fig:model}
\end{figure*}

Fig. \ref{fig:model} illustrates the architecture of our OpenFly-Agent, an aerial VLN model that builds upon the OpenVLA~\citep{openvla} baseline, since OpenVLA and aerial VLN share a similar pipeline, \emph{i.e.}, taking images and instructions as input and generating actions. OpenVLA is trained on 1M data, having strong abilities in instruction-following and reasoning, which establishes an efficient initialization for our model. In contrast, our OpenFly-Agent takes a sequence of images as input to indicate the observation history instead of one image in the original OpenVLA. Additionally, to mitigate visual redundancy between adjacent video frames while maintaining key information, two strategies are designed, \emph{i.e.,} keyframe selection and visual token merging. First, a series of candidate keyframes is selected based on the UAV flight trend and a landmark grounding module. Then, these keyframes are merged temporally, resulting in a compact sequence of visual tokens. Finally, the action decoder discretizes the predicted tokens to 6 action types specific to UAVs.

\subsection{Keyframe Selection}

The length of contextual visual tokens is a major challenge for VLMs when processing videos. Many open-source VLMs use uniform frame sampling \citep{buch2022revisiting, ranasinghe2024understanding, wang2025videoagent} to reduce calculation, but this strategy is not suitable for aerial VLN, since it may miss frames containing key landmarks. 
To address this issue, a keyframe selection strategy is proposed to emphasize important visual observations. We notice that sudden changes in the UAV's trajectory are often caused by the observation of landmarks, which can serve as a kind of cues to determine keyframes. Therefore, a heuristic method is adopted to select candidate frames by identifying the change point of the UAV's movement, followed by extracting the corresponding frame and two frames before and after it from the trajectory, constituting a keyframe set. Moreover, we design a landmark grounding module, which consists of three cross-attention layers to incorporate text and image features from the LLM hidden state, predicting the bounding boxes $\mathrm{b}\in\mathcal{R}^4$ of the instruction-indicated landmark. To incorporate as many landmark-related regions as possible into the historical visual tokens, candidate frames with the bounding boxes' area greater than the threshold $\theta$ will be retained as the final keyframes. During the training process, we obtain the bounding box of each landmark using the developed tools introduced in Sec. \ref{toolchain}, enabling the training of the grounding module and the accurate selection of keyframes. During the testing process, the bounding box of each frame is sequentially estimated by the well-trained grounding module. Then, our model selects keyframes by bounding boxes area and adjacent frames when a significant motion change occurs, forming a keyframe set for this moment.

\subsection{Visual Token Pruning}
To further reduce redundant information in keyframes, we introduce visual token merging into OpenFly-Agent. For the keyframes selected by the above method, a visual encoder maps them to multiple visual tokens, with each token representing the information of an image patch. Considering the potential inter-frame patch redundancy, we take a strategy that similar tokens in adjacent frames are periodically merged. Specifically, we select the frame with the largest bounding box in a keyframe set as the reference, since it usually contains the crucial observation indicating the landmark in an instruction. Then, we densely calculate the cosine similarities between each pair of visual tokens of the reference image and other comparative images in a keyframe set. 
Next, we merge the tokens with high similarity by averaging them, with the unmerged tokens in the comparative frame being discarded.
The merging operation is iteratively performed until the entire keyframe set has been traversed. Besides, we maintain a memory bank with a capacity of $K$ images, following a first-in-first-out (FIFO) policy to retain the latest keyframes. Since aerial VLN requires UAVs to perform long-distance flights based on instructions, we continue to conduct token compression within each keyframe to reduce the computational burden. The compressed visual tokens are obtained through grid pooling~\citep{llama_vid}. Notably, we keep the visual tokens of the current frame uncompressed to capture the latest visual observation, as it contains the most important information for action prediction.


%% file: sec/5_Experiments.tex
\section{Experiments}
\subsection{Implementation and Training Details}
\label{implementation}

The proposed OpenFly-Agent adopts the OpenVLA~\citep{openvla} as the baseline, with the current frame during flight remaining 256 tokens and all historical keyframes compressed into 1 token. The capacity $K$ of the history memory bank is set to 2 in our experiment. For the action head, the last 256 tokens in the vocabulary are used as special tokens for action representation. Similar to~\citep{aerialVLN,CityNav}, 6 actions for UAVs are defined as $\{$Forward, Turn Left, Turn Right, Move Up, Move Down, Stop$\}$. The OpenFly-Agent is trained with a batch size of 64 and a learning rate of 2e-5. The grounding module is optimized with a GIoU loss function, and the threshold $\theta$ for keyframe selection is set to 0.25 times the size of the input image.


\subsection{Evaluation Metrics}
Four standard metrics in VLN tasks are adopted to evaluate different methods, \emph{i.e.,}  navigation error (NE), success rate (SR), oracle success rate (OSR), and success weighted by path length (SPL). NE measures the average deviation between the UAV's final stopping point and the ground-truth destination. SR calculates the proportion of successful tasks, where a task is considered successful if the UAV stops within 20 m of the target~\citep{aerialVLN}. Each environment provides corresponding point clouds that enable collision checking. If a collision occurs, the task is counted as a failure. In OSR, if any point on the trajectory is within 20 m of the target, the task can be considered successful. SPL calculates the success rate weighted by the ratio of the ground-truth path length to the actually-executed path length.

\subsection{Quantitative Results}

We evaluate the proposed OpenFly-Agent and multiple VLN methods on the test set, with quantitative results listed in Table~\ref{tab:results}, where Seq2Seq, CMA, and AerialVLN achieve limited success rates. In contrast, Navid~\citep{navid} and NaVila~\citep{navila} are two most recent VLN methods, obtaining better results and demonstrating the great potential of VLMs in aerial VLN. See-Point-Fly~\cite{hu2025seepointflylearningfree} is a zero-shot method, which is evaluated using GPT-4.1 as the agent and demonstrates reasonable robustness. Our OpenFly-Agent outperforms the comparison methods by a large margin, benefiting from the proposed strategies. While aerial VLN is an emerging and challenging task, and there is still much room for improvement.
The results on the test-unseen split indicate the generalization abilities of these methods. Similarly, our method achieves the best performance, exhibiting a certain degree of robustness. However, all methods are significantly degraded, indicating that more powerful models are urgently needed to be developed.


\begin{table}[t!]
\centering
\caption{Comparison results on the test set. `Random’ means randomly
selecting one action to execute until the ‘stop’ action is chosen. All models are retrained using our dataset.}
\label{tab:results}
\resizebox{\textwidth}{!}{ 
\begin{tabular}{lccccccccc}
\toprule
\multirow{2}{*}{Method} & \multicolumn{4}{c}{test-seen} & \multicolumn{4}{c}{test-unseen}\\ 
\cmidrule(lr){2-5} \cmidrule(lr){6-9} 
& NE$\downarrow$ & SR$\uparrow$ & OSR$\uparrow$ & SPL$\uparrow$
& NE$\downarrow$ & SR$\uparrow$ & OSR$\uparrow$ & SPL$\uparrow$ \\ \midrule 

Random & 242m & 0.7\% & 0.8\% & 0\% & 301m & 0.1\% & 0.1\% & 0\%\\
Seq2Seq~\citep{VLN-CE} & 205m & 2.9\% & 24.3\% & 2.6\%& 229m & 2.1\% & 20.6\% & 1.1\%\\
CMA~\citep{VLN-CE}& 161m & 5.4\% & 28.1\% & 4.8\% & 217m & 4.6\% & 24.4\% & 2.1\%\\
See-Point-Fly~\citep{hu2025seepointflylearningfree} & - & - & - & -  & 191m & 8.2\% & 12.7\% & 6.3\%  \\
AerialVLN~\citep{aerialVLN} & 139m & 7.5\% & 30.0\% & 6.8\% & 214m & 7.3\% & 28.1\% & 4.4\%\\
Navid~\citep{navid}  & 153m & 13.0\% & 38.2\% & 11.6\% & 210m & 10.8\% & 27.2\% & 5.0\%\\
NaVila~\citep{navila} & \underline{132m} & \underline{20.3\%} & \underline{53.5\%} & \underline{17.8\%} & \underline{202m} & \underline{14.7\%} & \underline{42.1\%} & \underline{9.6\%}\\
OpenFly-Agent (Ours) & \textbf{93m} & \textbf{34.3\%} & \textbf{64.3\%} & \textbf{24.9\%} & \textbf{154m} & \textbf{22.6\%} & \textbf{56.2\%}  & \textbf{19.1\%}\\
\bottomrule
\end{tabular}
}
\end{table}

\begin{figure}[t]
\centering
    \includegraphics[width=0.98\linewidth]{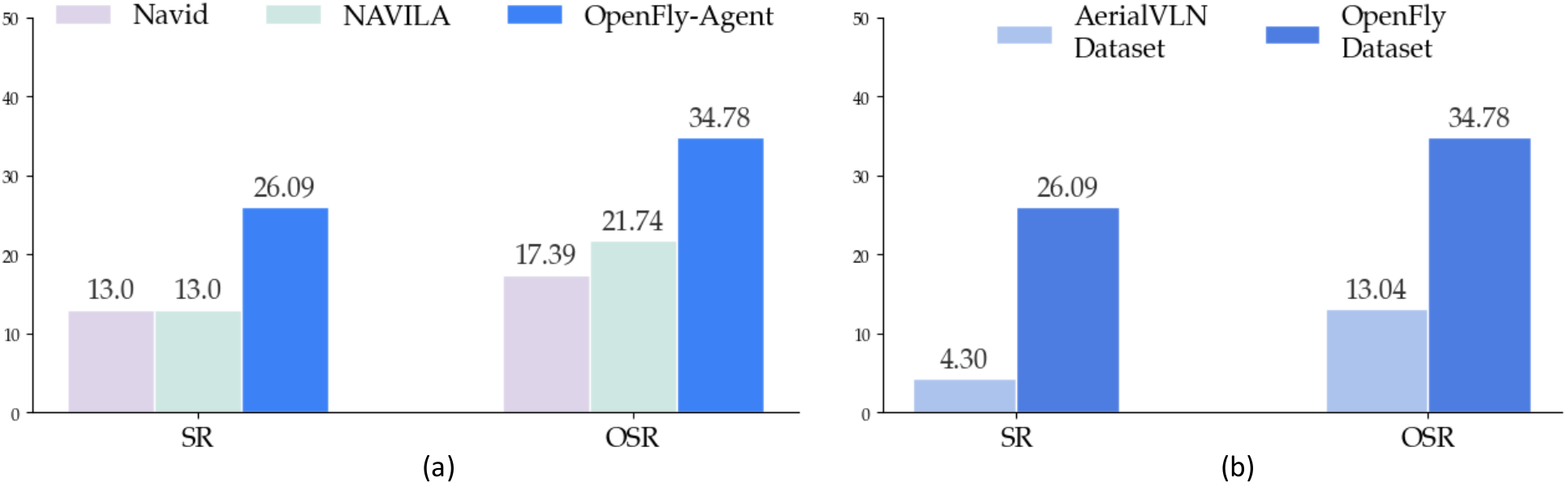}
    \caption{Results of real-world experiments. (a) Comparison with two strong VLN methods. (b) Performances of OpenFly-Agent trained on different datasets.}
    \label{fig:rr_data}
\end{figure}


\subsection{Real-world Experiments}
\label{appen_real_exp}

\begin{figure}[t]
\centering
\includegraphics[width=\linewidth]{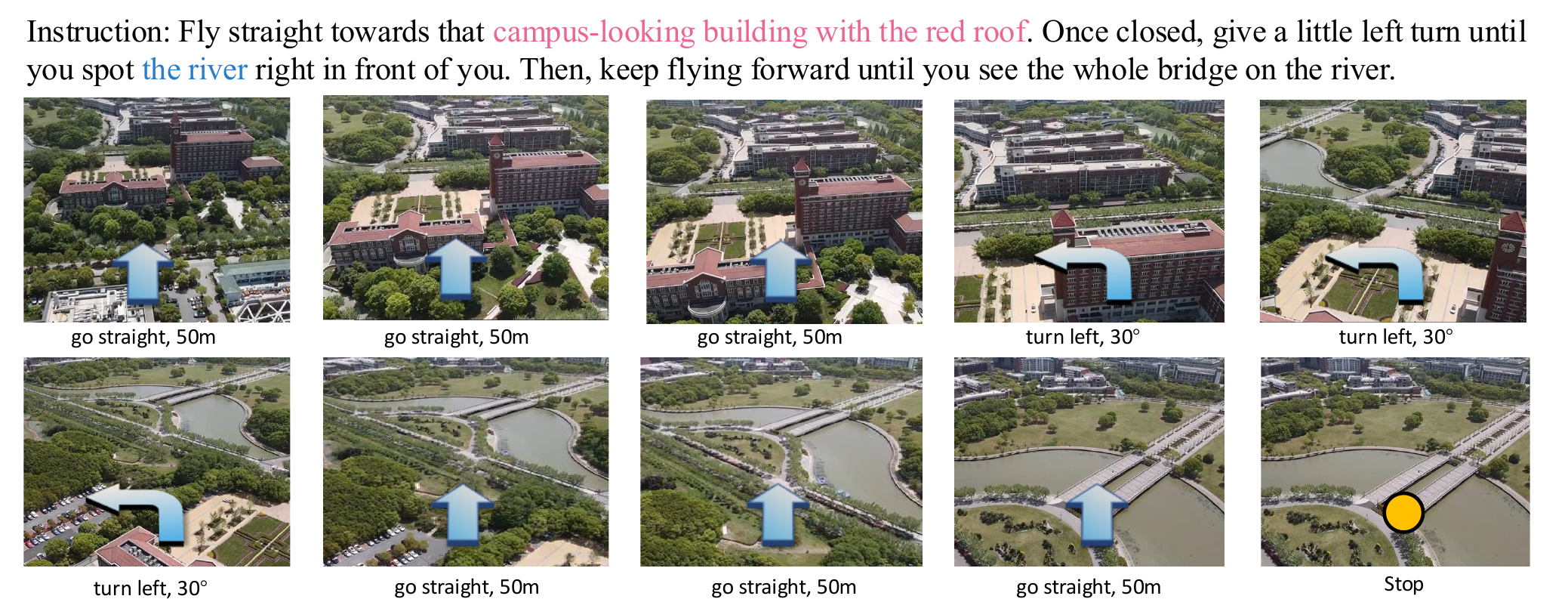}
    \caption{Snapshots of the real-world experiment.}
    \label{fig:real-world}
\end{figure}

The real-world experiments are conducted in 23 real outdoor scenes, where each scene corresponds to an unseen VLN task created by human operators, and the trajectory lengths range from 50m to 500m. We use a Q250 airframe as a real agent, carrying an NVIDIA Jetson Xavier NX running Ubuntu 18.04 as the onboard computer. In the real-world experiments with the drone, we utilize the “Super” ~\citep{ren2025safety} trajectory planning framework for local trajectory planning and employ Model Predictive Control (MPC)~\cite{falanga2018pampc} for trajectory tracking. The advantage of this paradigm is that it enables the VLN model to adapt to various planning and control algorithms, thereby accommodating diverse robotic platforms and scenarios. All methods run on an external PC communicating with the onboard computer to transfer images and action instructions. Two most recent models, Navid~\citep{navid} and NaVila~\citep{navila}, are evaluated for comparison. The results are shown in Fig~\ref{fig:rr_data} (a), where our model achieves the best performance with 26.09\% SR and 34.78\% OSR, significantly outperforming the comparison methods. This experiment again indicates the superiority of our OpenFly-Agent. Besides, we also trained our model on both our own dataset and the AerialVLN dataset separately. The results are shown in Fig.~\ref{fig:rr_data} (b), strongly demonstrating the capability of our data generation method in bridging the sim-to-real gap. A qualitative result is presented in Fig. \ref{fig:real-world}, and a dynamic demo can be found in our supplementary video.

\subsection{Ablation Study}
Ablation studies are conducted to evaluate the contribution of the keyframe selection and visual token merging in OpenFly-Agent. Table.~\ref{tab:ablation} shows the results, where OpenVLA~\citep{openvla} is our baseline. Using only the current frame or uniformly selecting from previous observation as keyframes makes the model perform poorly in the aerial VLN task. From `History + VTM', we can see that historical frames significantly improve the success rate. The keyframe selection strategy further increases the SR from 16.6\% to 34.3\%, demonstrating the effectiveness of key observations. Besides, the comparison between `KS' and `KS + VTM' indicates the great effect of our visual token merging strategy. We find that there is a severe imbalance between the number of text and image tokens if the token merging strategy is not applied. The cross-modal signal can be diluted by the numerical imbalance between a few text tokens and many visual tokens~\cite{univid}. As a result, background clutter, task-irrelevant distractors, and environmental noise may be encoded indiscriminately, leading to excessive computational cost and diluted attention to task critical cues~\cite{SemanticVLA}.

\begin{wraptable}[14]{r}{0.48\textwidth}
\centering
\vspace{-0.7cm}
\scriptsize
\caption{Ablation study on the test-seen split. `KS' and `VTM' denote keyframe selection and visual token merging, respectively. `History' indicates uniform frame sampling. `Random KS' means randomly selecting a frame from the candidate keyframe set.}
\label{tab:ablation}
\begin{tabular}{lcccc}
\toprule
Method& NE$\downarrow$ & SR$\uparrow$ & OSR$\uparrow$ & SPL$\uparrow$ \\ \midrule 

OpenVLA (baseline) & 231m & 2.3\% & 10.8\% & 2.2\% \\
History & 223m & 6.9\% & 23.3\% & 5.6\%\\
Random KS & 264m & 8.7\% & 26.6\% & 5.8\% \\
KS& 275m & 9.2\% & 28.1\% & 6.1\% \\
History + VTM & 215m & 16.6\% & 40.5\% & 9.1\%\\
KS + VTM & 93m & 34.3\% & 64.3\% & 24.9\%\\
\bottomrule
\end{tabular}
\end{wraptable}


%% file: sec/6_Conclusion.tex

\section{Conclusion}
In this work, we present OpenFly, a platform designed for large-scale data collection in aerial Vision-and-Language Navigation (VLN). OpenFly integrates multiple rendering engines and provides an automatic toolchain for data generation, enabling efficient collection of diverse, high-quality aerial VLN data. The resulting large-scale dataset comprises 100k trajectories across 18 distinct scenes, spanning a wide range of altitudes and lengths, which is significantly larger than existing ones. Furthermore, we propose OpenFly-Agent, a keyframe-aware aerial VLN model capable of identifying frames with critical observations, leading to accurate flight action prediction. Extensive experiments validate the effectiveness of the proposed method, and establish a comprehensive benchmark for future advancements in aerial VLN. 

%% file: sec/supp.tex
\section{More Details of Rendering Engines and Data Resources}
\label{appen_rendering_engines}

\begin{figure*}[h]
\begin{center}
   \includegraphics[width=\linewidth]{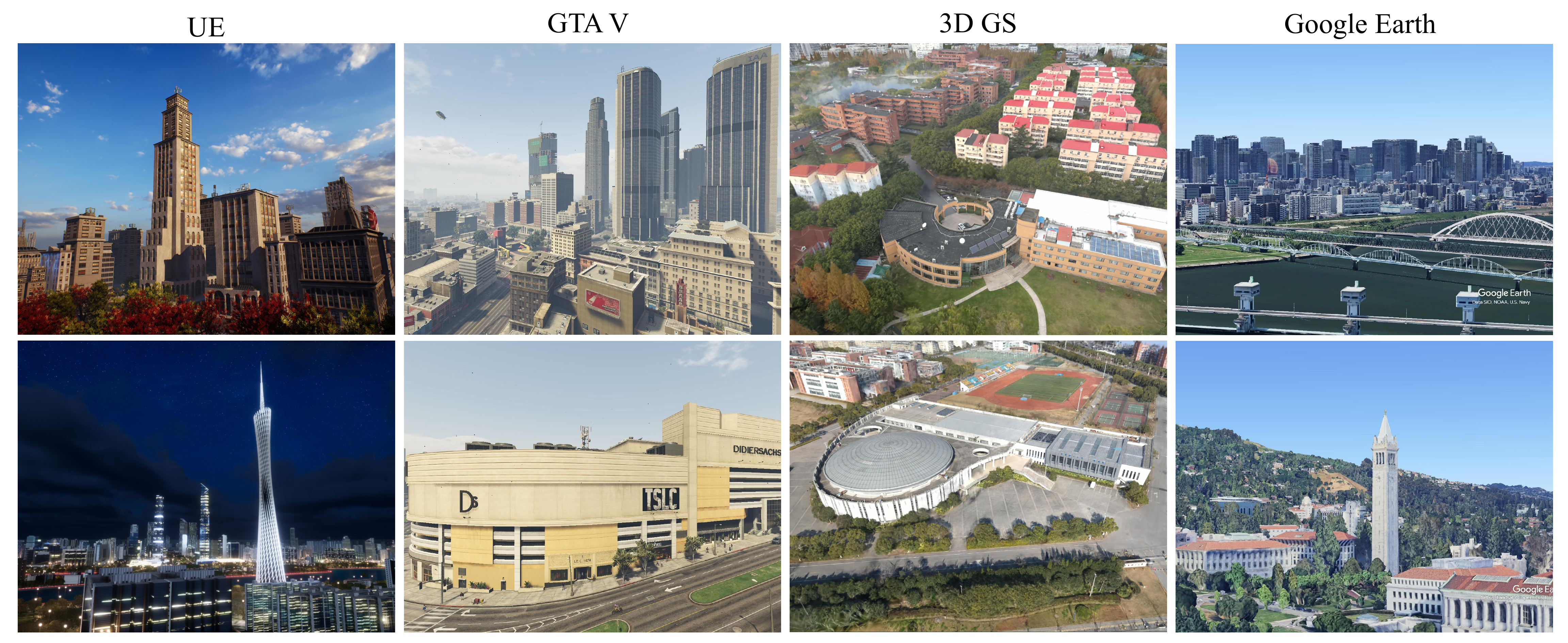}
\end{center}
   \caption{High-quality examples from different rendering engines and techniques, including several large cities such as Shanghai, Guangzhou, Los Angeles, Osaka, and etc., cover an area of over a hundred square kilometers in total. 3D GS provides five large campus scenes, further enhancing the diversity and realism of the data.}
\label{fig:all_dataset}
\end{figure*}

In this section, more information about the used rendering engines and data resources are detailed, with several high-quality examples illustrated in Fig. \ref{fig:all_dataset}. 

\textbf{Unreal Engine.} UE is a rendering engine capable of providing highly realistic interactive virtual environments. This platform has undergone five iterations, and each version features comprehensive and high-quality digital assets. In UE5, we meticulously select an official sample project named `City Sample', which provides us with a large urban scene covering $25.3 km^2$ and a smaller one covering $2.7 km^2$. These scenes include a variety of assets such as buildings, streets, traffic lights, vehicles, and pedestrians. Besides, in UE4, we prepare six more high-quality scenes. Specifically, there are two large scenes showcasing the central urban areas of Shanghai and Guangzhou, covering areas of $30.88 km^2$ and $58.56 km^2$, respectively. The remaining four scenes are selected from AerialVLN~\citep{aerialVLN}. They have smaller areas for totally about $26.64 km^2$. These scenes encompass a wide range of architectural styles, including both Chinese and Western influences, as well as classical and modern designs. Additionally, the UE4 engine allows us to make adjustments in scene time to achieve different appearances of scenes under varying lighting conditions.

AirSim is an open-source simulator, which provides highly realistic simulated environments for UAVs and cars. We integrate the AirSim plugin into UE4 to obtain image data easily from the perspective of a UAV. Since AirSim does not support UE5 and stopped updating in 2022, we use the UnrealCV~\citep{unrealcv} plugin as an alternative for image acquisition in UE5. To realize a highly efficient data collection in simulated scenes, we modify the UE5 project to a C++ project, integrate the UnrealCV plugin, and package executables for multiple systems like Windows and Linux. 

\textbf{GTA V.} 
It is an open-world game that is frequently used by computer vision researchers due to its highly realistic and dynamic virtual environment. The game features a meticulously crafted cityscape modeled after Los Angeles, encompassing various buildings and locations such as skyscrapers, gas stations, parks, and plazas, along with dynamic traffic flows and changes in lighting and shadows. 

Script Hook V is a third-party library with the interface to GTA V's native script functions. With the help of Script Hook V, we build an efficient and stable interface, which receives the pose information and returns accurate RGB images and lidar data. From the interface, we can control a virtual agent to collect the required data in an arbitrary pose and angle in the game.

\textbf{Google Earth.} 
It is a virtual globe software, which builds a 3D earth model by integrating satellite imagery, aerial photographs, and Geographic Information System (GIS) data. From this engine, we select four urban scenes covering a total area of $53.60 km^2 $, \emph{i.e.,} Berkeley, primarily consisting of traditional neighborhoods; Osaka, which features a mix of skyscrapers and historic buildings; and two areas with numerous landmarks: Washington, D.C., and St. Louis.

Google Earth Studio is a web-based animation and video production tool that allows us to create keyframes and set camera target points on the 2D and 3D maps of Google Earth. Using this functionality, we can quickly generate customized tour videos by selecting specific routes and angles. In order to efficiently plan the route, we develop a function that automatically draws the flight trajectory in Google Earth Studio according to the selected area and predefined photo interval. 

\textbf{3D Gaussian Splatting.} As a highly realistic reconstruction method, hierarchical 3D GS~~\citep{kerbl2024hierarchical} employs a hierarchical training and display architecture, making it particularly suitable for rendering large-scale areas. Due to these features, we use this method to reconstruct and render multiple real scenes. We utilize the DJI M30T drone as the data collection device, which offers an automated oblique photography mode, enabling us to capture a large area of real-world data with minimal manpower. Practically, we gathered data from five campuses across three universities, which are East China University of Science and Technology, Northwestern Polytechnical University, and Shanghai Jiao Tong University (referred to as ECUST, NWPU, and SJTU). These campus scenes include various types and styles of landmarks, such as libraries, bell towers, waterways, lakes, playgrounds, construction sites, and lawns. The detailed information for the five campuses is presented in Table~\ref{tab:GS_information}. 

SIBR~\citep{sibr2020} viewers is a rendering tool designed for the 3D GS project, enabling visualization of a scene from arbitrary viewpoints. The tool supports high-frame-rate scene rendering and provides various interactive modes for navigation. Building upon SIBR viewers, we developed an HTTP RESTful API that generates RGB images from arbitrary poses, simulating a UAV's perspective.

\begin{table}[t]
\caption{Different 3D GS Scenes}
\label{tab:GS_information}
\centering
\begin{tabular*}{\textwidth}{@{\extracolsep{\fill}}>{\centering\arraybackslash}p{0.5\textwidth} c c}
\toprule
Campus Name & Images & Area \\
\midrule
ECUST (Fengxian Campus) & 12008 & $1.06\,\text{km}^2$ \\
\midrule
NWPU (Youyi Campus) & 4648 & $0.8\,\text{km}^2$ \\
NWPU (Changan Campus) & 23798 & $2.6\,\text{km}^2$ \\
\midrule
SJTU (Minhang-East Zone) & 20934 & $1.72\,\text{km}^2$ \\
SJTU (Minhang-West Zone) & 9536 & $0.95\,\text{km}^2$ \\
\bottomrule
\end{tabular*}
\end{table}

\section{Details of 3D GS Data Collection}
From the UAV's perspective, choosing the appropriate shooting altitude poses a dilemma, \emph{i.e.,} if the altitude is too low, the sparse point cloud generated during the initialization of the 3D GS reconstruction will be suboptimal, due to insufficient feature point matches between photos. In contrast, if the altitude is too high, the Gaussian reconstruction will result in an overly coarse training of details. After multiple attempts, the data collection plan using the M30T was determined as follows. For large-scale block scenes, oblique photography is performed at approximately twice the average building height using the default parameters of the M30T’s wide-angle camera, with a tilt angle of -45°. For landmark buildings with heights significantly different from the average height, additional targeted data collection is conducted at twice their height. This altitude setting can, to a certain extent, ensure both higher-quality point cloud initialization and Gaussian splatting training.

\section{Unified Interfaces} 
\label{appen_interface}
For data collection, we integrate all rendering engines and design three unified interfaces, \emph{i.e.}, the agent movement interface, the lidar data acquisition interface, and the image acquisition interface, allowing an
agent to move and perceive the environment within any scene.
\begin{itemize}[left=0pt]
\item Agent Movement Interface: We design a \textit{CoorTrans} module, which implements a customized pose transformation matrix and scaling function to unify all coordinate systems into a meter-based FLU (Front-Left-Up) convention. This interface enables precise agent positioning, ensuring consistency and facilitating automatic trajectory generation.
\item Lidar Data Acquisition Interface: Lidar data is crucial for scene occupancy perception and essential for trajectory generation. Our platform supports different lidar data acquisition methods, including lidar sensor collection, depth map back-projection, and image feature matching. We develop a unified interface to integrate these methods and leverage the proposed \textit{CoorTrans} module to align all data to the same FLU coordinate system.
\item Image Acquisition Interface: We integrate HTTP RESTful and TCP/IP protocols to form a unified image request interface, allowing image data to be obtained from any location with flexible resolutions and viewpoints. 
\end{itemize}

\section{Results of Point Cloud Acquisition and Semantic Segmentation}
\label{appen_pc_ss}
\begin{figure}
    \centering
    \begin{subfigure}[b]{0.24\textwidth}
        \centering
        \includegraphics[width=\textwidth]{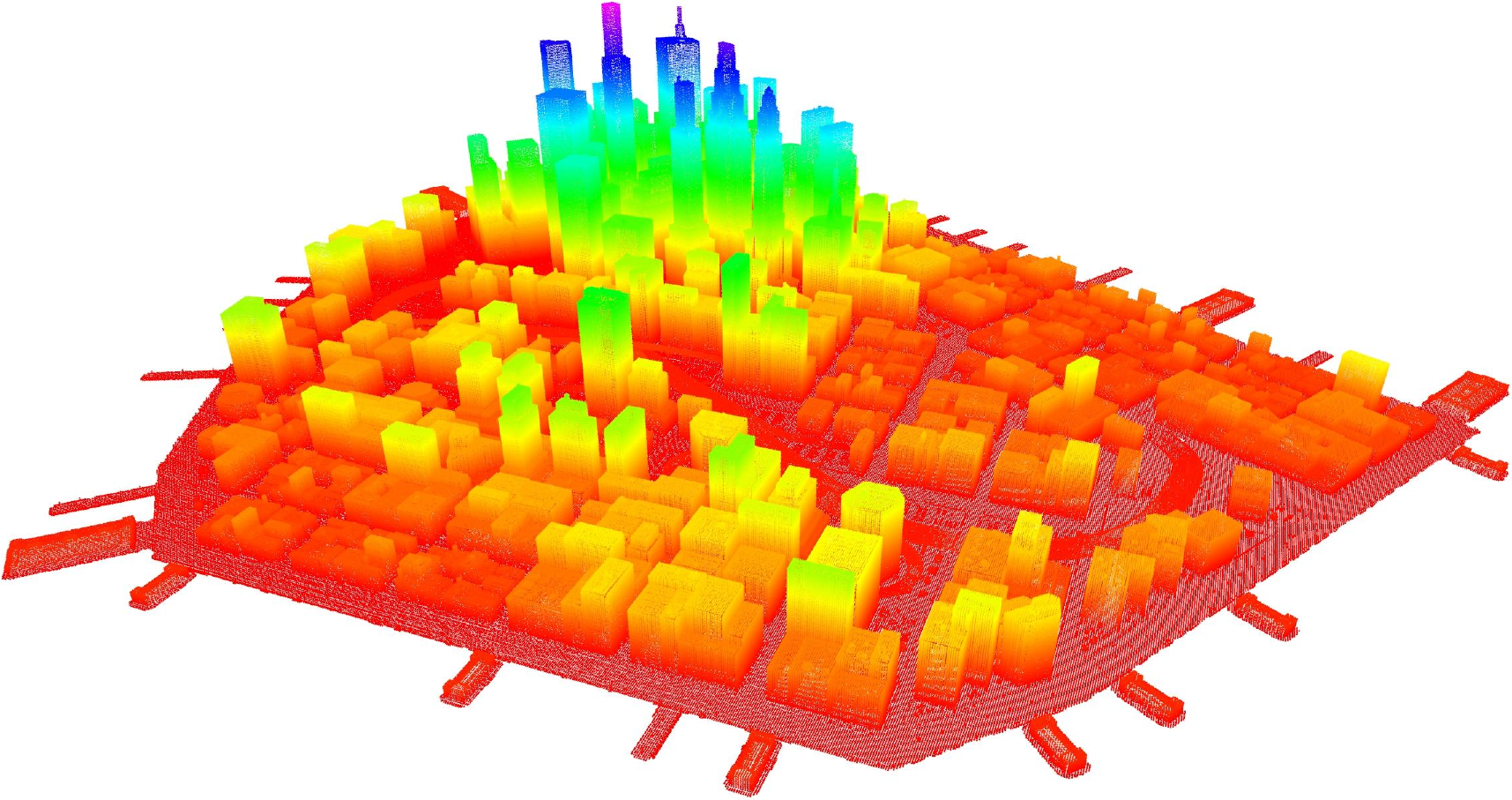}
        \caption{}
        \label{fig:pcd1}
    \end{subfigure}
    \hfill
    \begin{subfigure}[b]{0.2\textwidth}
        \centering
        \includegraphics[width=\textwidth]{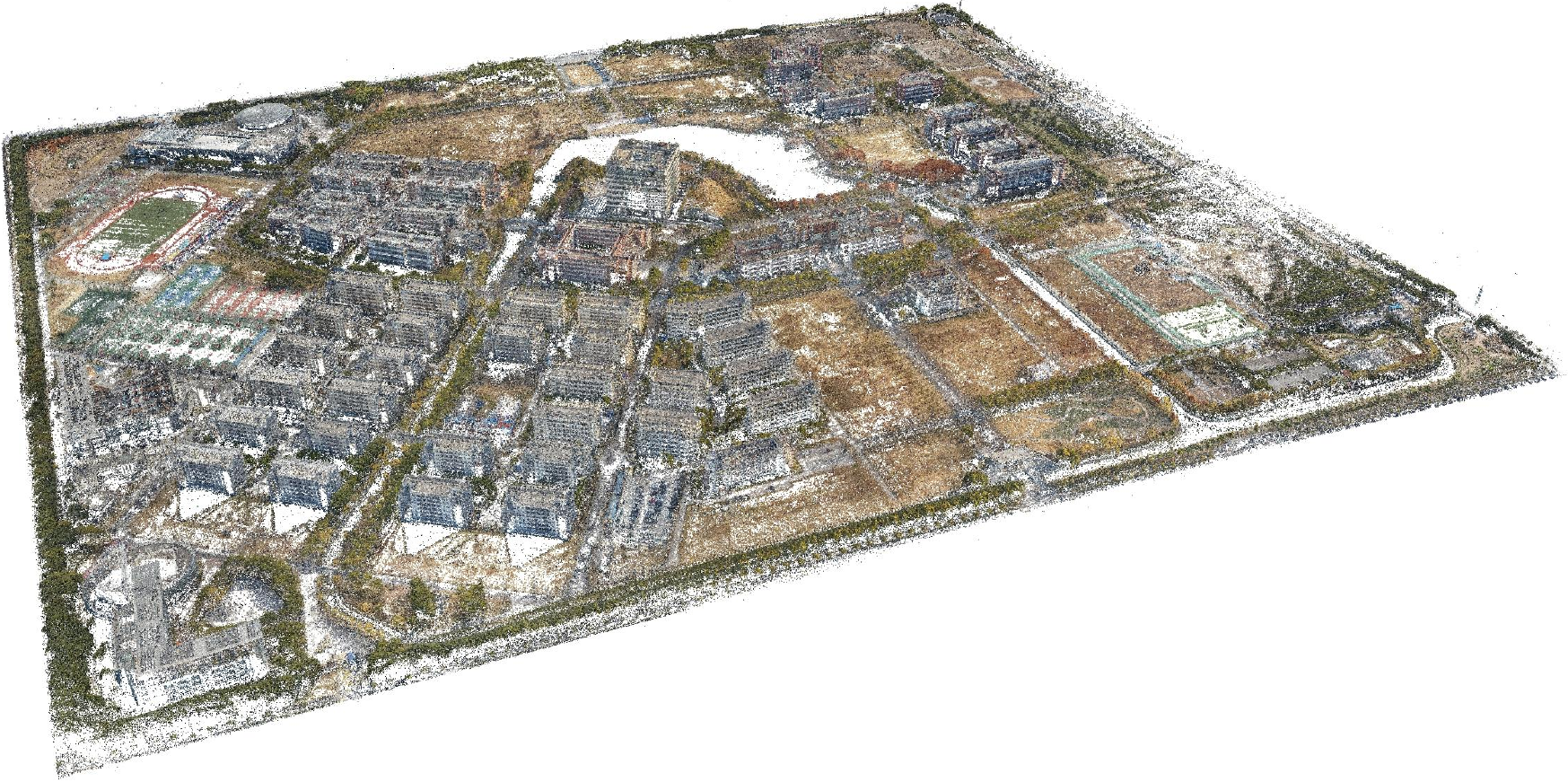}
        \caption{}
        \label{fig:pcd2}
    \end{subfigure}
   \hfill
    \begin{subfigure}[b]{0.24\textwidth}
        \centering
        \includegraphics[width=\textwidth]{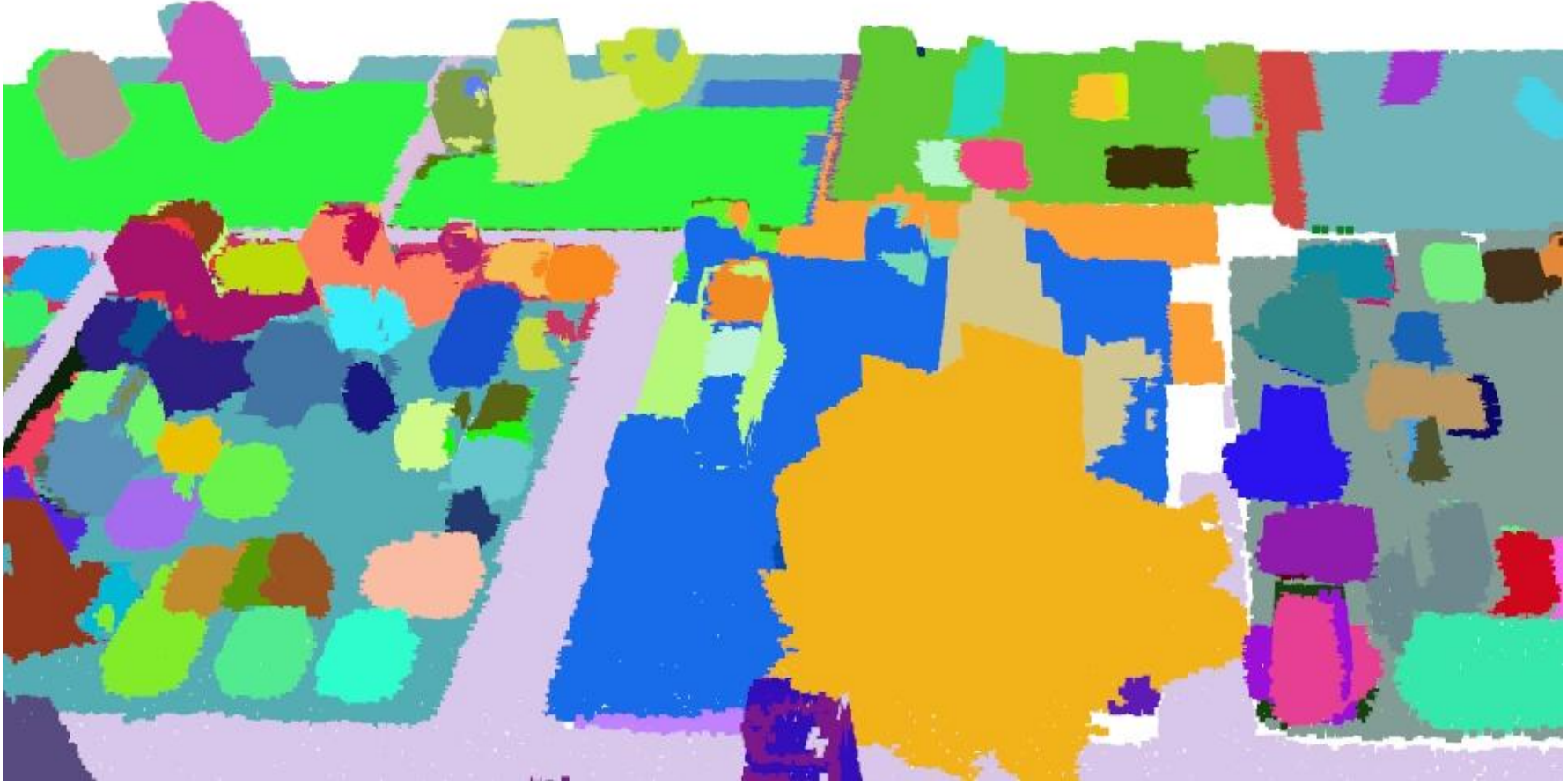}
        \caption{}
        \label{fig:seg2}
    \end{subfigure}
    \hfill
    \begin{subfigure}[b]{0.24\textwidth}
        \centering
        \includegraphics[width=\textwidth]{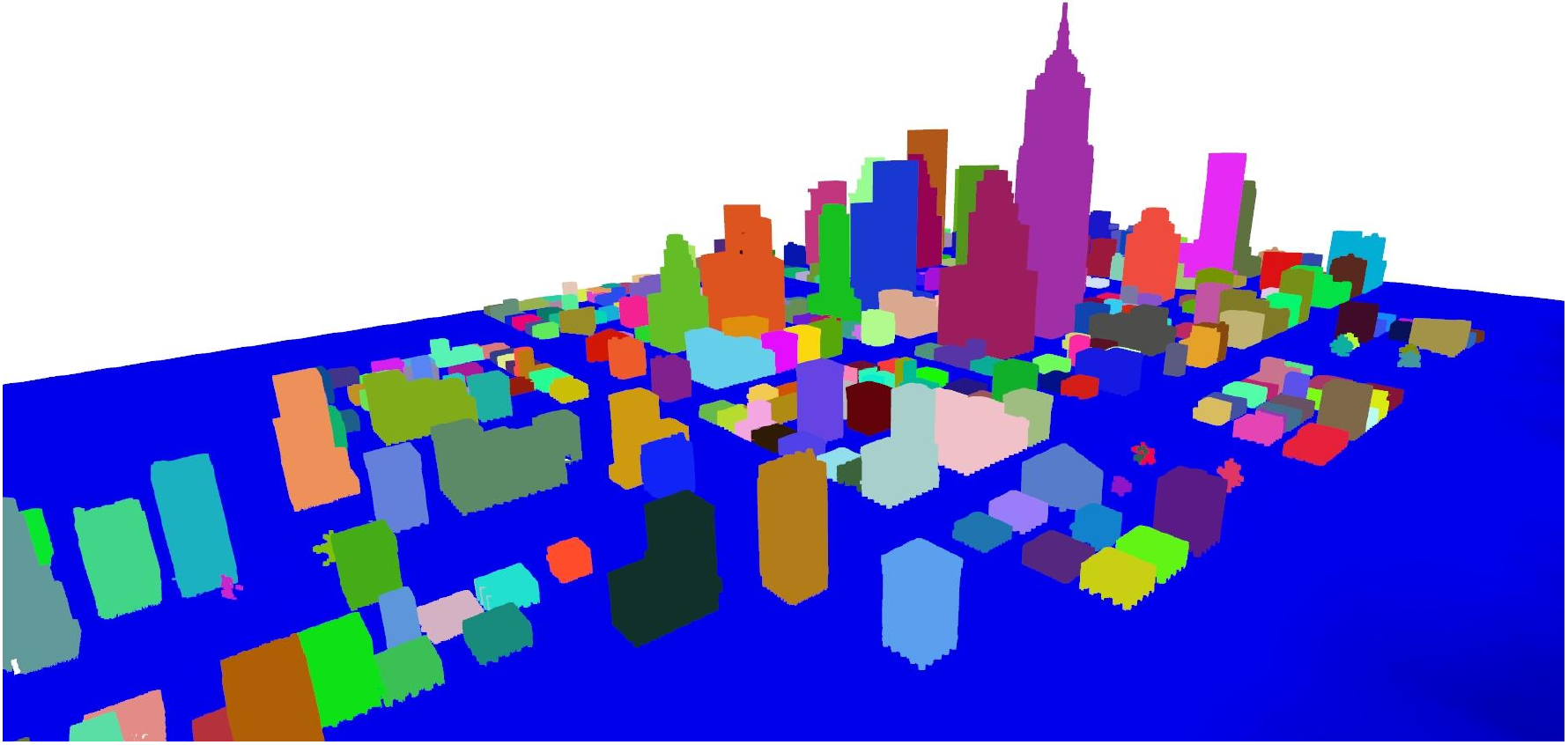}
        \caption{}
        \label{fig:seg1}
    \end{subfigure}

    \caption{Results of our point cloud acquisition and semantic segmentation. (a) Rasterized sampling point cloud reconstruction. (b) Image-based point cloud reconstruction. (c) Semantic 3D scene segmentation. (d) Point cloud projection and contour extraction.}
    \label{fig:pcd_seg_results}
\end{figure}

This section presents results regarding the point cloud acquisition and scene semantic segmentation, as shown in Fig. \ref{fig:pcd_seg_results}.

\textbf{Point Cloud Acquisition.} 
OpenFly integrates data resources from different rendering engines. In order to obtain point cloud information from different scenes, we provide two point cloud acquisition methods. 1) For UE and GTAV scenes, we provide a tool that utilizes rasterized point cloud sampling and reconstruction to obtain a global point cloud. The results can be seen in Fig.~\ref{fig:pcd1}. 2) For 3D GS scenes, we use COLMAP~\citep{colmap} to obtain relatively sparse point data from images, as shown in Fig.~\ref{fig:pcd2}. Although the point cloud generated by this method is relatively sparse, it provides sufficient coverage information.

\textbf{Scene Semantic Segmentation.} 
To meet the requirements of different data resources and different segmentation granularities, our OpenFly offers three methods for obtaining 3D semantic segmentation of scenes. Here, we present results of segmentation methods other than manual annotation. 1) Fig.~\ref{fig:seg2} illustrates the semantic segmentation results based on the off-the-shelf 3D scene understanding method Octree-Graph~\citep{octree_graph}. This method provides more granular results. 2) The result of semantic segmentation via point cloud projection and contour extraction is shown in Fig.~\ref{fig:seg1}. This method leverages high-precision point clouds to achieve instance segmentation for structures like buildings and trees, which directly contact the ground.

\section{A Comprehensive Introduction to Automatic Trajectory Generation}
\label{appen_trajectory}
Leveraging the point cloud map and segmentation tools, OpenFly offers two methods for trajectory generation for different scenes. 
1) Path search based on customized action space: First, a global voxel map $M_{global}$ and a bird's eye view (BEV) occupancy map $M_{bev}$ are constructed from the scene point cloud. Second, the flight altitude is randomly selected within the user-defined height range, and landmarks that are not lower than the height threshold $H_{\tau}$ are chosen as targets. A starting point is selected within the distance range $[r, R]$ from the landmark, ensuring that it is not occupied in both $M_{global}$ and $M_{bev}$. Then, a point on the line connecting the starting point and the landmark, which is close to the landmark and unoccupied in $M_{bev}$, is chosen as the endpoint. 
Third, A collision-free trajectory from the starting point to the endpoint is generated using the A*~~\citep{astar} pathfinding algorithm, where the granularity of exploration step size and direction can be adjusted according to the action space. Besides, by repeatedly selecting the endpoint as the new starting point, complex trajectories can be generated. Finally, utilizing OpenFly's interface, images corresponding to the trajectory points can be obtained. 2) Path search based on grid: Google Map data does not allow image retrieval at arbitrary poses in the space. Thus, we rasterize a pre-selected area and collect images from each grid point in all possible orientations. Starting and ending points are chosen within the grid points to generate trajectories. Corresponding images for these trajectory points are then selected from the pre-collected image set.

\section{Details of Instruction Generation}
\label{appen_instruction}

Except for the instruction generation described in the main paper. The remaining process is mainly divided into two parts: landmark feature extraction and sub-instruction fusion. A simplified prompt to the VLM and the corresponding response are probably like this.

\begin{itemize}[left=0pt]
    \item Get Landmark features. \\
    \textbf{System Prompt}: You are an assistant who is proficient in image recognition. You can accurately identify the object in the picture and its characteristics that are different from the surrounding objects. I will give you the three final images you will see. Please focus on the last image and tell me the features of the target building and reply to me in the form of JSON.

    \textbf{User}: The target is the nearest prominent landmark to me. Answer me a dictionary like color:--, feature: --, size: --, type: --.

    \textbf{GPT 4o}: color: blue, feature: Steel, glass, size: medium size, type: building.

    \item Instruction Fusion. \\
    \textbf{System Prompt}: You are an assistant proficient in text processing. You need to help me combine these scattered actions and landmarks into a sentence using words with similar meanings and more appropriate words, making them smooth, fluent, and accurate. If the landmarks of adjacent actions are similar or even identical, please use pronouns to refer to them.

    \textbf{User}: Multiple sub-instructions.

    \textbf{GPT 4o}: Move forward to a high-rise building with a noticeable logo at the top. Then, slightly turn left and go straight to a futuristic tower with a large spherical structure in the middle.
\end{itemize}









\section{Data Quality Control}
\label{appen_quality}
\textbf{Data Filter.}
During data collection, it is inevitable that some damaged or low-quality data will be generated. We clean the data in the following situations. 1) We remove damaged images that are produced in generation or transmission. 2) We find that UAVs sometimes appear to pass through the tree models. Therefore, we exclude the trajectories where the altitude is lower than that of the trees. 3) We believe that extremely short or long trajectories are not conducive to model training. Thus, we remove these trajectories, specifically those with fewer than 2 or more than 150 actions.

\textbf{Instruction Refinement.}
A known challenge of instruction generation is VLMs' hallucinations. During the previous instruction generation process, sometimes the same landmark appears across several frames. This results in a VLM generating similar captions for the repeated observations of a landmark, increasing the complexity of the final instruction and introducing ambiguity due to duplication.

To mitigate this challenge, we utilize the NLTK library ~\citep{bird2006nltk} to simplify the instruction by detecting and merging similar descriptions. Specifically, a syntactic parse tree is first constructed to extract all landmark captions using a rule-based approach. Then, a sentence-transformer model is employed to encode the extracted landmark captions into embedding vectors. Their similarities are computed with dot product, and high-similarity captions are then identified and replaced with referential pronouns (\emph{e.g.}, ``it," ``there," \emph{etc.}). For example, a generated instruction with redundant information is ``$\cdots$ make a left turn toward \textbf{a medium-sized beige building marked by a signboard reading CHARLIE'S CHOCOLATE}. Continue heading straight, passing \textbf{a medium-sized gray building with a prominent rooftop billboard displaying Charlie’s Chocolate} $\cdots$". After simplification, a more concise sentence is obtained, \emph{i.e.,} ``$\cdots$ make a left turn toward \textbf{a medium-sized beige building marked by a signboard reading CHARLIE'S CHOCOLATE}. Continue heading straight, passing \textbf{it} $\cdots$", demonstrating the effectiveness of this post-processing technique. 

\textbf{Manually Check.}
At the same time, we built a data inspection platform to provide instructions, action sequences, and corresponding images to human examiners. If an instruction describes all the actions and landmarks in a trajectory well, it is considered qualified. We randomly select 3K samples from the entire dataset according to the data distribution. After manually inspecting these samples, we find that they reach a high qualification rate of 91\%. There is some ambiguity in the description of some landmarks in the remaining data, making it likely that these landmarks are not easily distinguishable from the surrounding environment. However, the examiners consider this not entirely unacceptable. In summary, most of the generated data feature good quality for the aerial VLN task.

\section{More Dataset Analyses}
\label{appen_dataset_analysis}

\begin{wrapfigure}{r}{0.5\textwidth}
    \vspace{-1cm}
    \centering
    \includegraphics[width=0.48\textwidth]{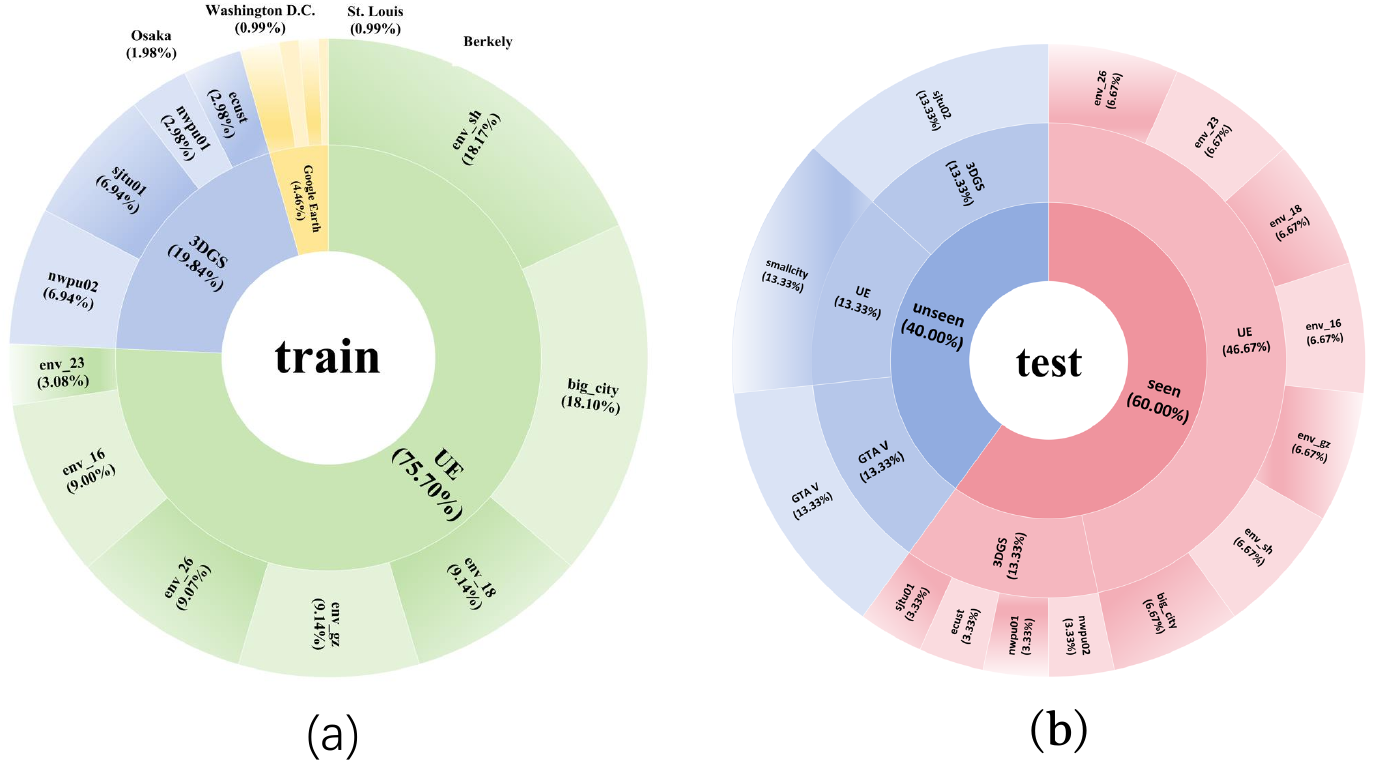} 
    \caption{The distribution of the data volume in different scenes under the Train and Test sets. (a) Train set distribution. (b) Test set distribution.}
    \label{fig:traj_sta}
\end{wrapfigure}

Following previous studies~\citep{aerialVLN, wang2025, ANDH}, we conducted a statistical analysis of the linguistic phenomena using 25 randomly selected instructions and compared the results with other VLN datasets, as detailed in Table~\ref{tab:linguistic_phenomena}. The analysis shows that the generated instructions exhibit rich linguistic phenomena such as `Reference' and `Comparison'. Notably, our dataset is not the most complex one, since we believe that instructions in VLN tasks should be more aligned with real-life scenarios, rather than emphasizing length and complexity. This cognition is consistent with that of REVERIE~\citep{REVERIE}. Our instructions avoid overly lengthy and unrealistic expressions to some extent, making them more practical to command UAVs.

Fig.~\ref{fig:traj_sta} (a) shows the data distribution of the train set, where 7 UE scenes account for 75.7\% of the total 100K data, 4 3D GS scenes account for nearly 20\% of the total amount, and Google Earth data accounts for 4.46\%. Fig.~\ref{fig:traj_sta} (b) presents the data distribution of the test set, where the seen data and unseen data account for 60\% and 40\%, respectively.

\begin{table*}[!t]
    \centering
    \caption{Linguistic phenomena analysis using randomly selected 25 instructions. $p$ denotes the proportion of instructions showing the phenomenon, and $\mu$ represents the average number of times the phenomenon occurs in each instruction.}
    \resizebox{\linewidth}{1.4cm}{
        \begin{tabular}{l| c c| c c| c c| c c|c}\hline
                & \multicolumn{2}{c|}{R2R} & \multicolumn{2}{c|}{ANDH} & \multicolumn{2}{c|}{AerialVLN}& \multicolumn{2}{c|}{OpenFly} & \\
            Phenomenon & $p$ & $\mu$ & $p$ & $\mu$ & $p$ & $\mu$ & $p$ & $\mu$ & Example in OpenFly\\ \hline
            
            Reference & 100 &   3.7 &   92  &   1.9       &    100 &   9.7  &  100 &  2.7  & ...Advance to \textbf{the large beige and brown building with windows}... \\
            
            Coreference &   32  &   0.5 &   8   &   0.1  &   68  &   1.8 &   52  &   1.4&{...Continue moving forward to reach  \textbf{it} ...}\\
            
            Comparison &    4   &   0.0 &   32   &   0.4 &    20  &  0.2   &   60 &   0.7  &    ...Move ahead to the \textbf{medium-sized} beige building...\\ 
            
            Sequencing &    16  &   0.2 &   8  &   0.1   &   68 &   3.7 &   64 &   0.8& ...move forward to  \textbf{next} large light brown building ...\\
            
            Allocentric Relation &  20  &   0.2 &   32  &   0.4 &   56  &   4.6  &   76 &   1.0   &...black large building featuring a billboard \textbf{on} its rooftop... \\
            
            Egocentric Relation &   80  &   1.2 &   32  &   0.4  &   100 &   7.1  &   100  &   2.4  &... then slightly turn left as you move ahead towards... \\
            
            Imperative &    100 &   4.0 &   100&   1.1   &  100 &   6.9 &   100 &   3.6   &   {.... Proceed slightly straight and turn left ....} \\
            
            Direction & 100 &   2.8 &   100 &   1.4     &  100 &   4.6   &   100  &   3.8 & ... \textbf{turn right and go ahead} to ... \\
            
            Temporal Condition &    28  &   0.4 &   20  &   0.2  &   76  &   5.6 &   72 &   0.9  &   ...Continue straight \textbf{until} you reach ... \\\hline
        \end{tabular}
    }
    \label{tab:linguistic_phenomena}
   
\end{table*}

\section{Qualitative Experimental Results}
\label{appen_qualitative}
\begin{figure}[!ht]
\centering
    \includegraphics[width=\linewidth]{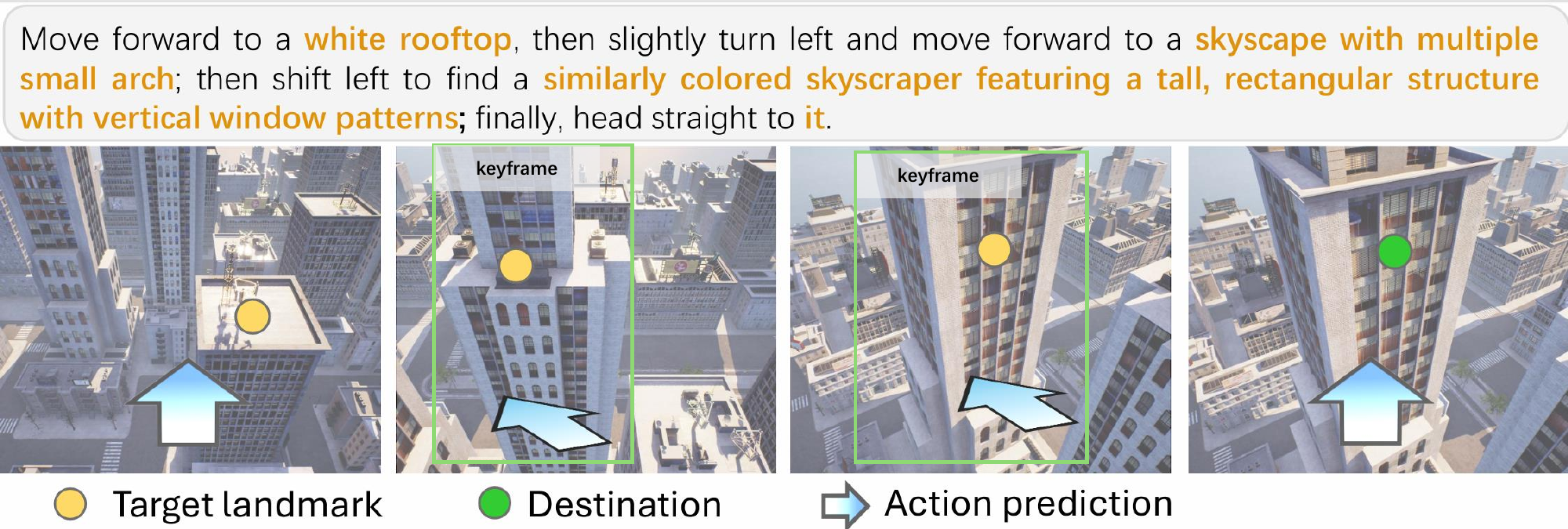}

    \caption{Illustration of an aerial VLN trajectory generated by OpenFly-Agent in a UE scene, which successfully predicts actions following the instruction when encountering landmarks. The green bounding box represents the correct landmark prediction.}
    \label{fig:example}
\end{figure}

\begin{figure}[!h]
\centering
    \includegraphics[width=\linewidth]{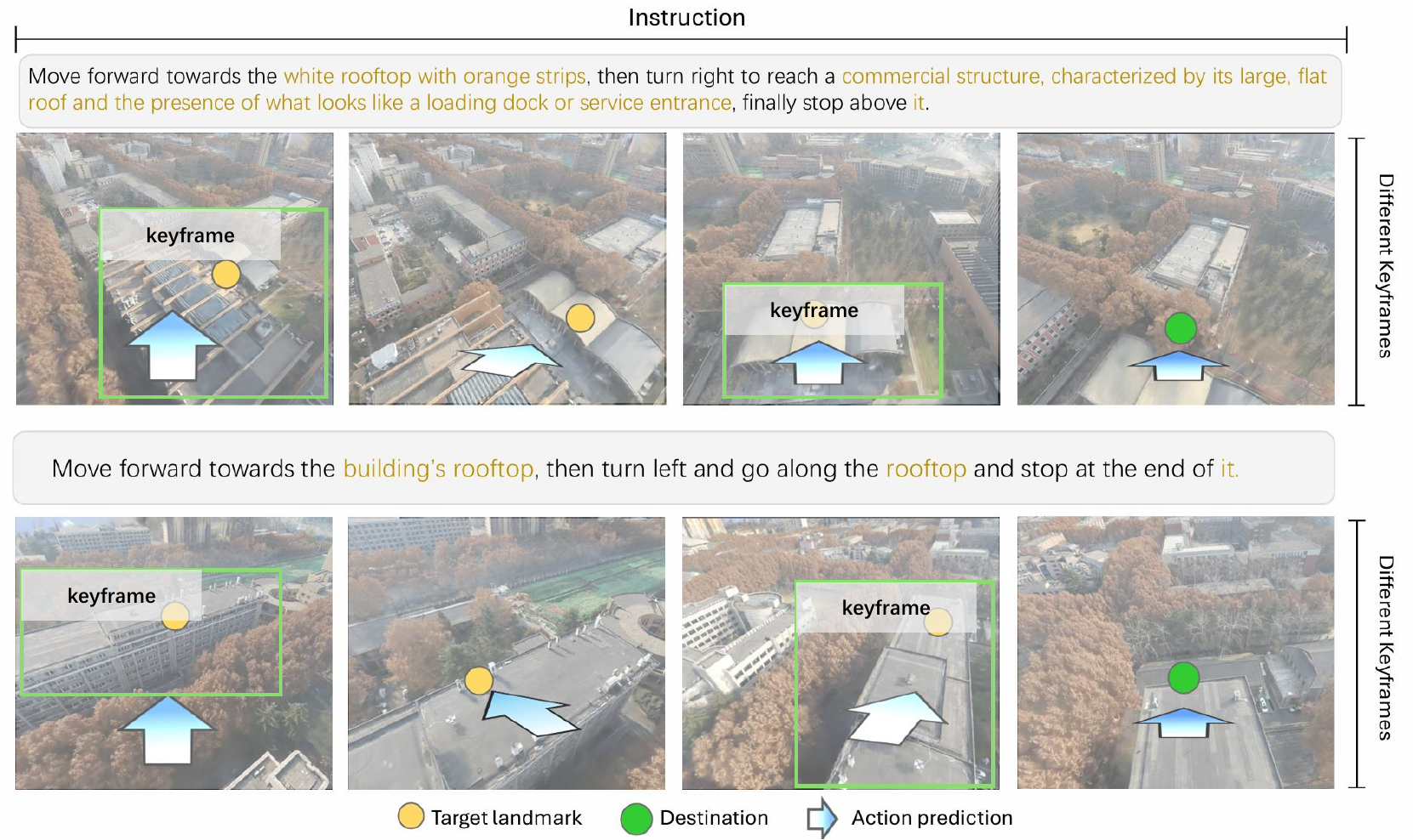}
    \caption{Illustration of aerial VLN trajectories generated by OpenFly-Agent in a 3D GS scene. The green bounding box represents the correct landmark locations prediction.}
    \label{fig:success_case}
\end{figure}

\begin{figure}[!tp]
\centering
\includegraphics[width=0.7\linewidth]{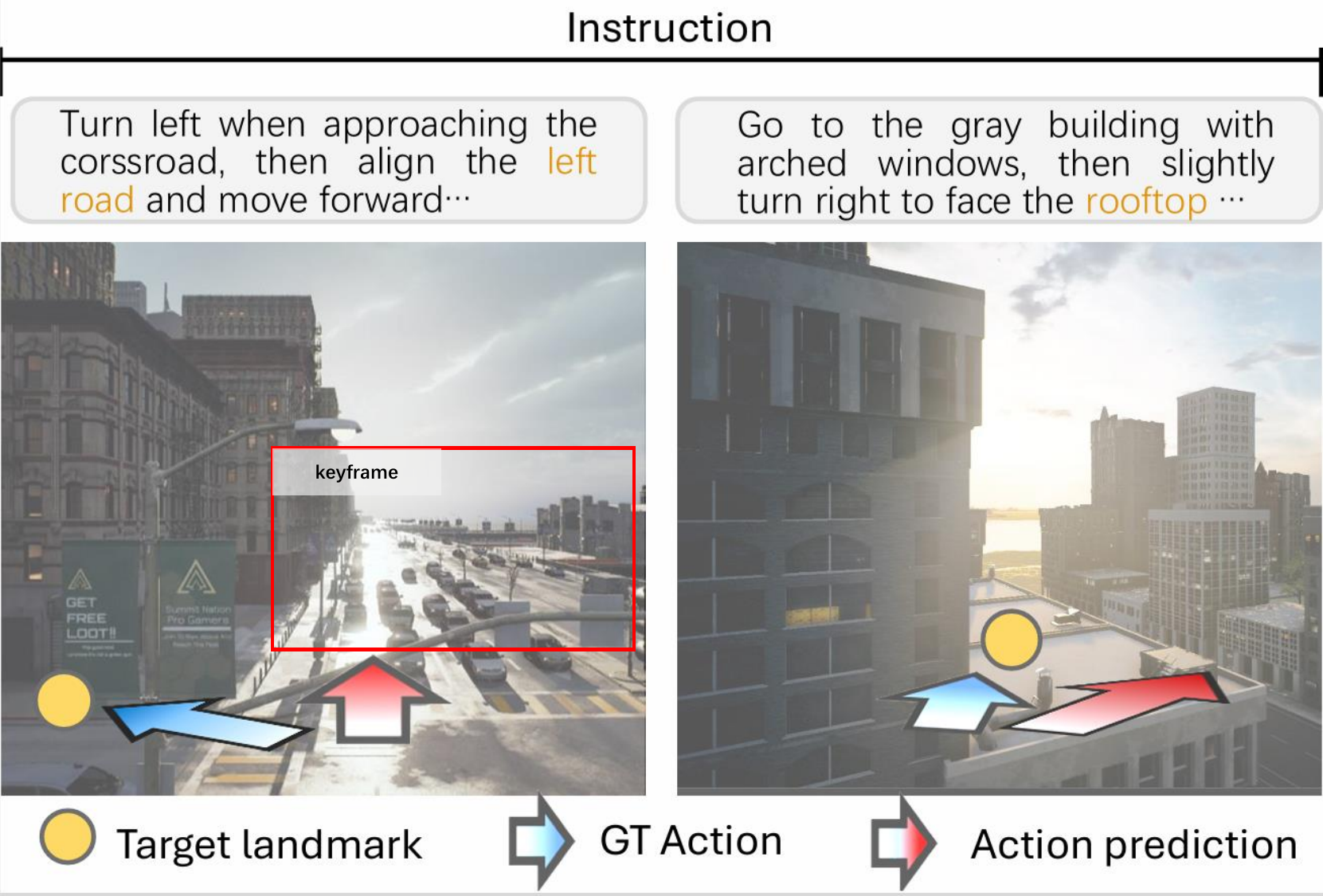}
    \caption{Illustration of failure cases. Sometimes our model may misclassify key landmarks or output wrong actions. The red bounding box represents incorrect landmark locations.}
    \label{fig:fail}
\end{figure}

\begin{figure}[!tp]
\centering
\includegraphics[width=0.7\linewidth]{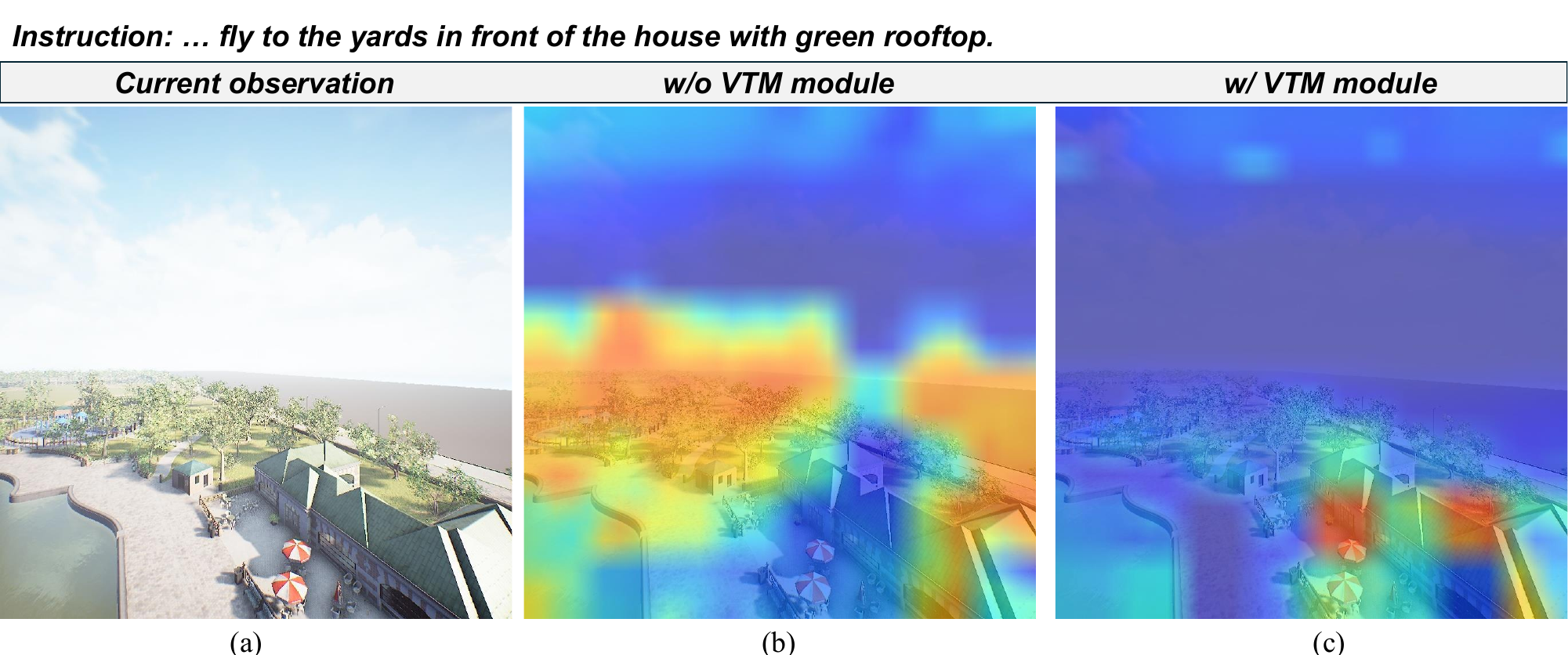}
    \caption{Visualization of attention maps of current observation patches. a) Current observation. b) The attention map without the VTM module and c) the attention map with the VTM module, reflecting token merging strategy matters in avoiding diluting attention of current observation.}
    \label{fig:vtm_case}
\end{figure}

Fig.~\ref{fig:example} presents a qualitative result in a UE scene, where our OpenFly-Agent successfully navigates to the destination according to the instruction. It presents a powerful capability in perceiving environments and aligning observations with complex instructions. Fig.~\ref{fig:success_case} presents another successful aerial VLN example in a 3D GS scene. The image style, flight heights, and viewpoints are significantly different from UE's scenarios. In this case, our OpenFly-Agent exhibits robustness to handle data with great diversity. In addition, Fig.~\ref{fig:fail} shows two failure cases, where our model sometimes fails to identify the landmark or output actions with proper amplitudes. To further verify the superiority of the `VTM' module, we illustrate the attention maps with and without `VTM' in Fig.~\ref{fig:vtm_case}, demonstrating its effect in avoiding diluting the attention of the current observation.

\clearpage

\section{Use of LLMs}
In this work, we employ large language models (LLMs) to automatically identify and correct grammatical errors, thereby improving the overall fluency and readability of the generated text.